\documentclass[sigconf]{acmart}
\AtBeginDocument{%
  \providecommand\BibTeX{{%
    \normalfont B\kern-0.5em{\scshape i\kern-0.25em b}\kern-0.8em\TeX}}}

\setcopyright{acmcopyright}
\copyrightyear{2018}
\acmYear{2018}
\acmDOI{10.1145/1122445.1122456}

\acmConference[Woodstock '18]{Woodstock '18: ACM Symposium on Neural
  Gaze Detection}{June 03--05, 2018}{Woodstock, NY}
\acmBooktitle{Woodstock '18: ACM Symposium on Neural Gaze Detection,
  June 03--05, 2018, Woodstock, NY}
\acmPrice{15.00}
\acmISBN{978-1-4503-XXXX-X/18/06}

\usepackage{microtype}
\usepackage{graphicx}
\usepackage{color}
\usepackage{mathrsfs}
\usepackage{cleveref}
\usepackage{float}
\usepackage{subfigure}
\usepackage{diagbox}
\usepackage{booktabs}
\usepackage{makecell}
\usepackage{multirow}
\usepackage{CJKutf8}
\begin{document}

\title{GlyphCRM: Bidirectional Encoder Representation for Chinese Character with its Glyph}

\author{Yunxin Li}

\affiliation{%
  \institution{Harbin Institute of Technology, Shenzhen}
  \country{China}
}
\email{liyunxin987@163.com}

\author{Yu Zhao}
\affiliation{%
  \institution{Harbin Institute of Technology, Shenzhen}
  \country{China}
  }
\email{zhaoyuhitsz@163.com}

\author{Baotian Hu}
\affiliation{%
  \institution{Harbin Institute of Technology, Shenzhen}
  \country{China}
}
\email{hubaotian@hit.edu.cn}

\author{Qingcai Chen}
\affiliation{%
 \institution{Harbin Institute of Technology, Shenzhen}
 \country{China}
}
\author{Yang Xiang}
\affiliation{%
 \institution{Peng Cheng Laboratory, China}
 \country{China}
}
\email{xiangy@pcl.ac.cn}

\author{Xiaolong Wang}
\affiliation{%
 \institution{Harbin Institute of Technology, Shenzhen}
 \country{China}
}
\author{Yuxin Ding}
\affiliation{%
 \institution{Harbin Institute of Technology, Shenzhen}
 \country{China}
}
\author{Lin Ma}
\affiliation{%
 \institution{Meituan, Beijing}
 \country{China}
}
\email{forest.linma@gmail.com}

\renewcommand{\shortauthors}{Yunxin Li and Yu Zhao, et al.}

\begin{abstract}
Previous works indicate that the glyph of Chinese characters contains rich semantic information and has the potential to enhance the representation of Chinese characters. The typical method to utilize the glyph features is by incorporating them into the character embedding space. Inspired by previous methods, we innovatively propose a Chinese pre-trained representation model named as GlyphCRM, which abandons the ID-based character embedding method yet solely based on sequential character images. We render each character into a binary grayscale image and design two-channel position feature maps for it. Formally, we first design a two-layer residual convolutional neural network, namely HanGlyph to generate the initial glyph representation of Chinese characters, and subsequently adopt multiple bidirectional encoder Transformer blocks as the superstructure to capture the context-sensitive information. Meanwhile, we feed the glyph features extracted from each layer of the HanGlyph module into the underlying Transformer blocks by skip-connection method to fully exploit the glyph features of Chinese characters. As the HanGlyph module can obtain a sufficient glyph representation of any Chinese character, the long-standing out-of-vocabulary problem could be effectively solved. Extensive experimental results indicate that GlyphCRM substantially outperforms the previous BERT-based state-of-the-art model on 9 fine-tuning tasks, and it has strong transferability and generalization on specialized fields and low-resource tasks. We hope this work could spark further research beyond the realms of well-established representation of Chinese texts.

\end{abstract}





\maketitle
\section{Introduction}

\begin{figure}[t]
    \centering
    \includegraphics[width=0.45\textwidth]{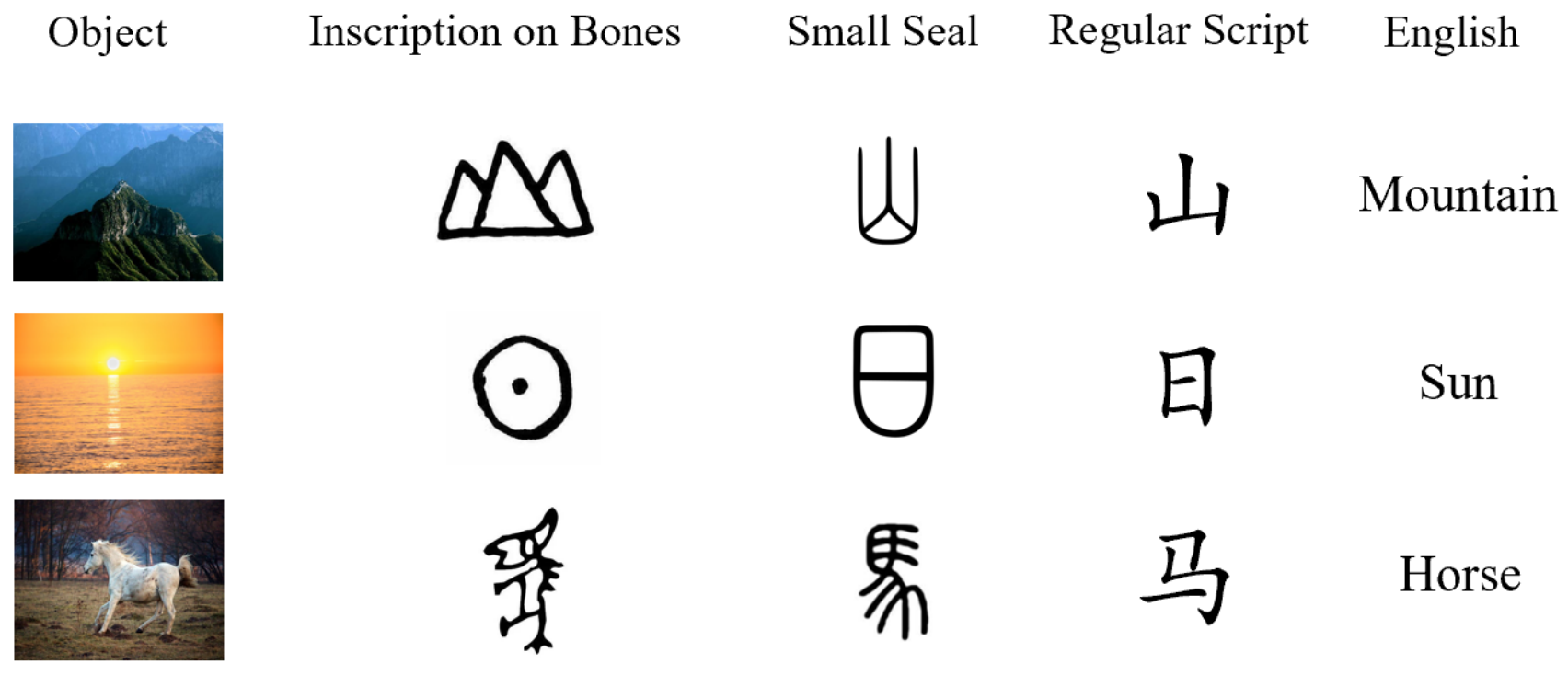}
    \caption{An example of the evolution of some Chinese characters. Their glyphs are similar to the physical objects in nature and contain rich semantic information.}
    \label{fig:charater}
\end{figure}

\begin{CJK}{UTF8}{gbsn}
Pre-trained neural language models, e.g., BERT~\cite{devlin-etal-2019-bert}, BART~\cite{lewis-etal-2020-bart}, and XLNET~\cite{xlnet}, have achieved extraordinary success in many natural language processing~(NLP) tasks, such as Information Retrieval~\cite{ir_bert}, Semantic Matching~\cite{semantic_bert,reimers-gurevych-2019-sentence_bert}, Question Answering~\cite{wang-etal-2017-questiona, questiona2} and Text Classification~\cite{text_classification}. In these models, each word is usually converted into a discrete vector representation by looking up the word embedding table, and then the context-sensitive word representation is learned by certain structures. However, for Chinese texts, these routine methods ignore that the static glyph of Chinese characters contains rich semantic information. For instance, Pictographs: the shape of '山'~(mountain), '日'~(sun) and '马'~(horse) is inextricably related to the shape of natural objects as shown in Figure~\ref{fig:charater}; Radicals: '崎'~(rough) and '岖'~(rugged), which have the same radical '山'~(mountain), are usually used to describe things related to mountains~(e.g., mountain road) together. Hence, the glyphs of Chinese characters can convey some meanings in many cases, and Chinese characters with similar structures can have intrinsic links. They intuitively indicate that the glyph features of Chinese characters have the potential to enhance their representations.
\end{CJK}

\begin{figure}[t]
    \centering
    \includegraphics[width=0.45\textwidth]{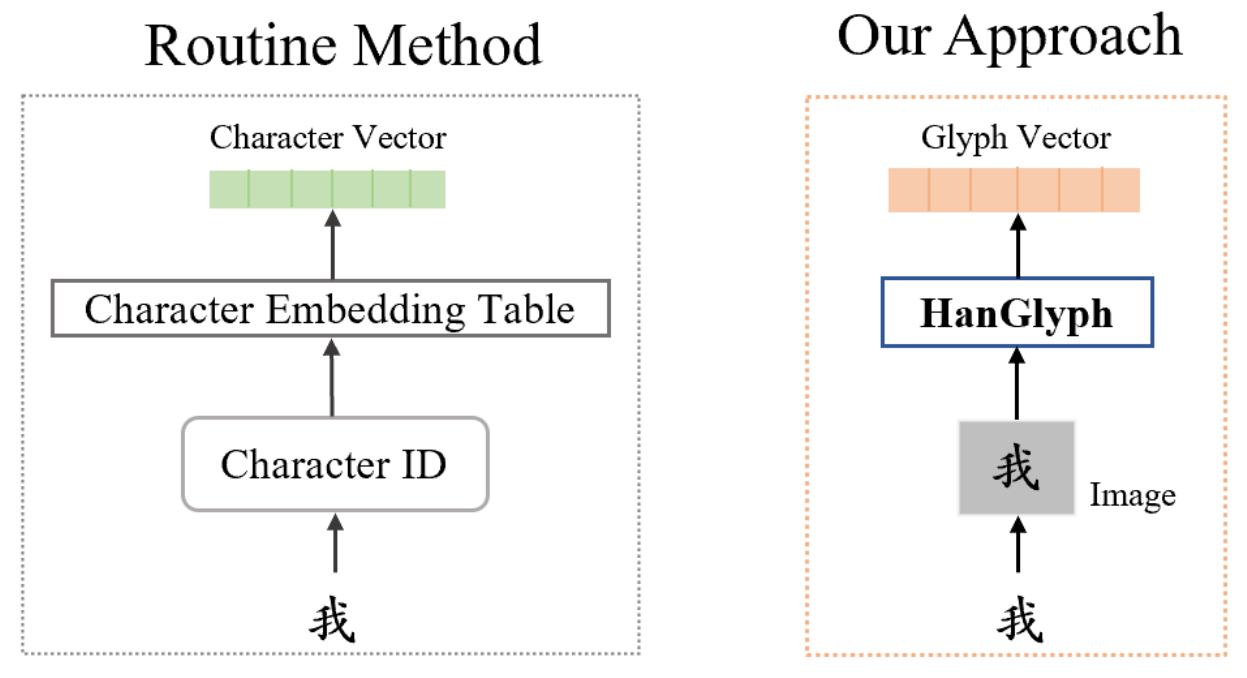}
    \caption{Left: The routine representation method of Chinese characters, Right: Our approach based exclusively on glyphs of Chinese characters.}
    \label{fig:charater_2}
\end{figure}

Based on the above observations, some methods~\cite{learning_glyph, glyce, chen-etal-2020-glyph2vec} incorporate the glyph features to enhance the Chinese character representation already covered into character embeddings~(character ID-based), e.g., \citet{glyce} combine the glyph features extracted from various forms of Chinese characters with the BERT embeddings. They demonstrate that the glyph features of Chinese characters are authentically helpful to improve the performance of models. However, previous methods only use glyphs of Chinese characters as the additional features, and there is no pre-trained Chinese text representation framework based on glyphs. In this paper, as shown in Figure~\ref{fig:charater_2}, instead of using the ID-based character embedding method, we propose to only use the glyph vectors of Chinese characters as the their representations, obtained by the HanGlyph module. To capture the contextual information, we further adopt the bidirectional encoder Transformer~\cite{attention} as the superstructure and finally propose the Chinese pre-trained representation model named GlyphCRM, based entirely on glyphs.



Concretely, we design two residual convolutional blocks in the HanGlyph module to obtain the glyph representation of any Chinese character, which is converted into the grayscale image. Each block has similar sub-layer architectures, including convolutional neural networks (CNN) and ReLU~\cite{relu} activation function. Furthermore, we design two-channel position maps for the character image to reinforce the capture of the spatial structure of Chinese characters' glyphs. Meanwhile, to fully exploit the glyph features of Chinese characters, we incorporate the glyph features extracted by the HanGlyph module into the underlying two Transformer blocks by the skip-connection method. From the whole architecture of GlyphCRM, it does not use the ID-based word/character embedding method and can be fine-tuned for specific NLU tasks, where Chinese characters only need to be converted into grayscale images. Hence, the proposed model can address the out-of-vocabulary problem and alleviate the issue of huge parameters stemming from the word/character embedding table. 

When pre-training GlyphCRM, we adopt the Masked Language Model (Masked LM) pre-training object and Next Sentence Prediction (NSP) task, identical to BERT~\cite{devlin-etal-2019-bert}. Furthermore, we evaluate models on 9 down-stream Chinese NLU tasks, including text classification, sentence matching, and tagging. Extensive experimental results demonstrate that GlyphCRM significantly outperforms the previous state-of-the-art pre-trained representation model BERT on a wide range of Chinese tasks. The in-depth analysis indicates that it converges faster than BERT during pre-training and has strong transferability and generalization on specialized fields and low-resource tasks.      

The contributions of our paper are three-fold:
\begin{itemize}
    \item We propose a Chinese pre-trained representation model Gly-phCRM based entirely on glyphs for the first time, where it replaces the ID-based word/character embedding method with the convolutional representation of Chinese glyphs. 
    \item GlyphCRM addresses the out-of-vocabulary problem by the HanGlyph module, which can generate the glyph representation of any character already converted into the grayscale image when fine-tuned on specific tasks.
    \item Extensive experiments demonstrate that our proposed model achieves better performance on a wide range of Chinese NLU tasks, especially on sequence labeling, compared to BERT with similar Transformer depth.
\end{itemize}

\section{Related Work}


\textbf{Pre-trained Language Representation Models:} As the distributional representation of words is proved more efficient and practical than independent representation that ignores the contextual information~\cite{sahlgren2008distributional}, how to obtain the rich context-sensitive representation of words has been attracting promising attention of many researchers. Early methods such as Word2Vec~\cite{word2vec} and GloVe~\cite{pennington-etal-2014-glove} learn the word embeddings with fixed dimensions through the co-occurrence of words in fixed windows on large-scale corpora. Recently, to alleviate the problem of insufficient representation of the above methods, some researchers study how to learn the word embeddings that contain more comprehensive contextual information and long-distance dependency information between words. \citet{elmo, elmo_b} proposed ELMo and its successors that utilize the language models to capture the contextual features with left-to-right and right-to-left methods. 

As the simple yet efficient Transformer~\cite{attention} architecture emerg-es, recently proposed pre-trained language models adopt it as the main architecture and have achieved significant performances on many NLU tasks. For instance, GPT and its successors~\cite{gpt,radford2019language_gpt_2, gpt_3} utilize the Transformer decoder-based architecture with the self-supervised left-to-right pre-training method to obtain the context-sensitive representation of words. Different from GPT, BERT~\cite{devlin-etal-2019-bert} utilizes the self-attention mechanism-based Transformer architecture and adopts the Masked Language Model pre-training object and Next Sentence Prediction task to obtain the bidirectional representation of words. Moreover, sequence-to-sequence pre-trained models such as T5 \cite{T5} and BART~\cite{lewis-etal-2020-bart}, and BERT-based representation models such as RoBERTa~\cite{liu2019roberta}, XLNET~\cite{xlnet}, DeBERT \cite{debert}, also achieve promising gains on NLP tasks.

Many researchers also proposed pre-trained language models for different languages such as CamemBERT for French~\cite{martin-etal-2020-french_bert}, BERT-wwm~\cite{scir_wwm} and ERNIE~\cite{zhang-etal-2019-ernie} for Chinese. Multi-language pre-trained models such as mT5~\cite{mt5} are also proposed to handle the compounded language tasks such as machine translation expediently. To summarize, the language models pre-trained on large-scale corpora are extremely useful for the development of natural language processing. In this paper, we also adopt the popular directional encoder Transformer as the superstructure of our model, which uses the cross attention mechanism to capture the context-sensitive information. We pre-train it on large-scale Chinese corpora.

\begin{figure}[t]
    \centering
    \includegraphics[width=0.43\textwidth]{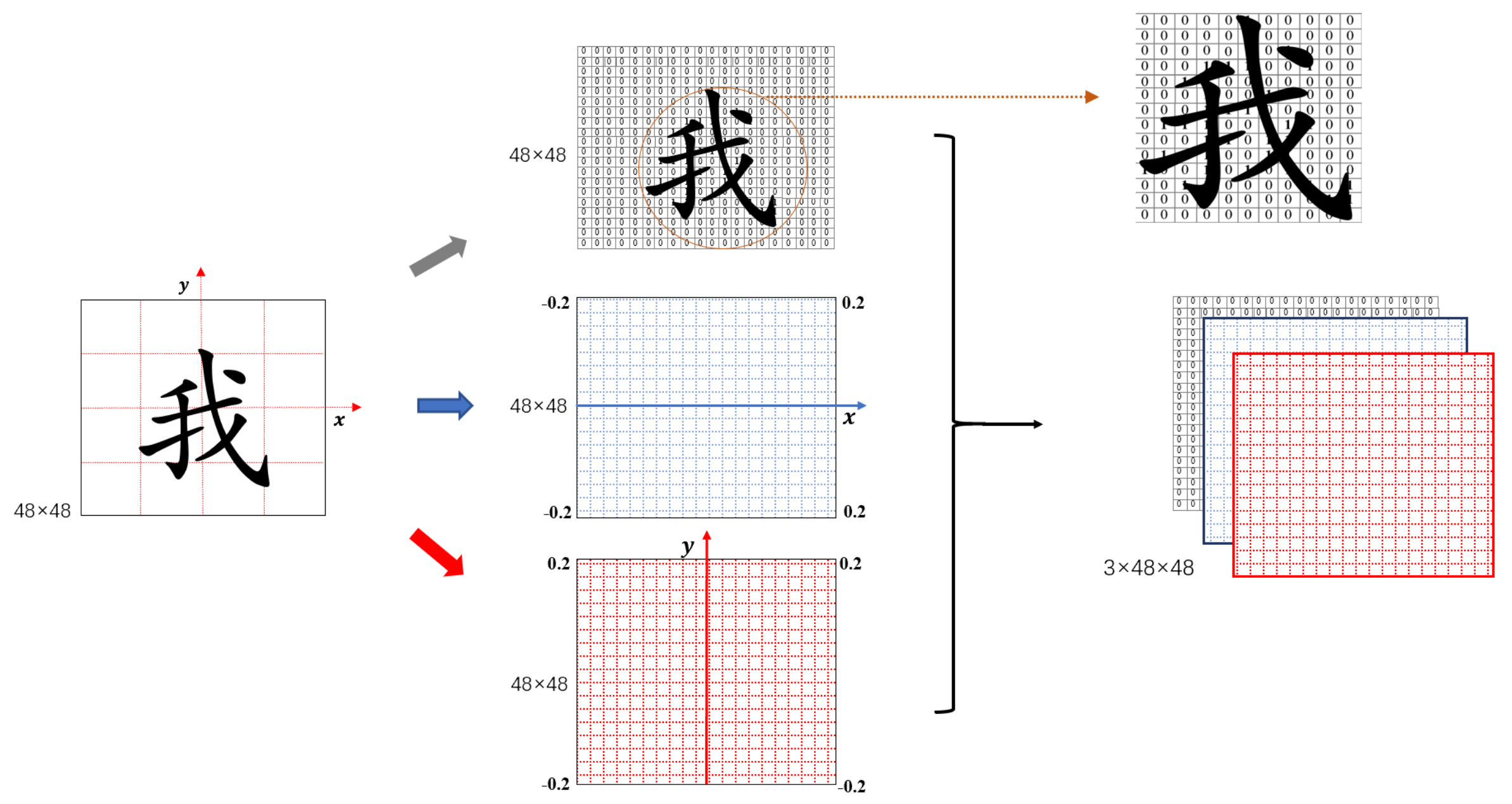}
    \caption{\begin{CJK}{UTF8}{gbsn}An preprocessing instance of Chinese character '我' (me). Each column of abscissa position map (the second one) has the same value. Each row of ordinate position map (the third one) has the same value.\end{CJK}}
    \label{fig:axis}
\end{figure}
\noindent\textbf{Glyph Vector:} Generally, to represent the discrete symbolic texts, researchers proposed to encode various symbols into the corresponding word embedding space, such as one-hot encoding and distributed representation, making a remarkable progress in the field of natural language processing. However, from the perspective of symbolic evolution, Chinese symbols have always possessed their unique structural features and peculiarities, developing from the initial hieroglyphics to their present forms as the Chinese characters shown in Figure \ref{fig:charater}. Recent researches~\cite{learning_glyph, chen-etal-2020-glyph2vec} also demonstrate that the glyphs of Chinese characters contain rich semantic information and have the potential to enhance the word representation of them. \citet{glyce} first apply the glyph features of Chinese characters into the pre-trained model BERT and achieve significant performance on many Chinese NLU tasks, such as Named Entity Recognition~\cite{msra}, News Text Classification~\cite{thunews} and Sentiment Analysis~\cite{chnSentiCorp}. Among them, the typical method is to use the deep convolutional neural networks to extract the glyph features of Chinese characters after converting them into images. Then, the glyph features of Chinese characters and corresponding character embeddings are integrated to enrich the representation of Chinese characters. We argue that the full glyph of Chinese characters is expressive enough, and further propose the Chinese pre-trained representation model GlyphCRM, based entirely on glyphs.

\section{Our Methodology}

In what follows, we first introduce the data preprocessing and then mainly introduce the overall architecture of GlyphCRM, presented in Figure~\ref{fig:model}, containing the HanGlyph module and bidirectional encoder layer. Finally, we introduce the two-stage pre-training and fine-tuning methods of our model in detail.

\subsection{Data Preprocessing}
For the input text, we render each Chinese character into a single-channel $ 48\times48$ grayscale image\footnote{We use Python's fontTools and PIL library to set the font of Chinese characters and render Chinese characters into images.}. As the instance shown in Figure \ref{fig:axis}, the position on each character feature map~(i.e. grayscale image) is set to '1' where the stroke of Chinese character passes; otherwise, the position is set to '0'. After obtaining the sequential feature maps of character-level input text, we apply a special token $\rm[CLS]$ as the start symbol for all inputs. We also apply another special token $\rm[SEP]$ to separate sentences and as the ending symbol of input for pre-training and specific NLU tasks. The two special tokens are also converted to the grayscale image to obtain the corresponding feature maps. Hence, the input and output sequence can be separately denoted to $\mathbf{X} = (x_{cls}, x_1, x_2, ..., x_n, x_{sep})$, and $\mathbf{H} = (\mathbf{h}_{cls}, \mathbf{h}_1, \mathbf{h}_2, ..., \mathbf{h}_n, \mathbf{h}_{sep})$.

To further capture the spatial structure of Chinese characters, we design the identical two-channel position maps for each character image, which have the same size as the feature maps of Chinese characters. As shown in Figure~\ref{fig:axis}, we set the coordinate axis with the center point of the Chinese character image as the origin, and the value range of the horizontal and vertical axis is between $-0.2$ and $0.2$. Hence, the two-channel position maps separately represent the abscissa and ordinate values of each pixel after being projected, respectively.

\subsection{HanGlyph}
\begin{figure}[t]
    \centering
    \includegraphics[width=0.45\textwidth]{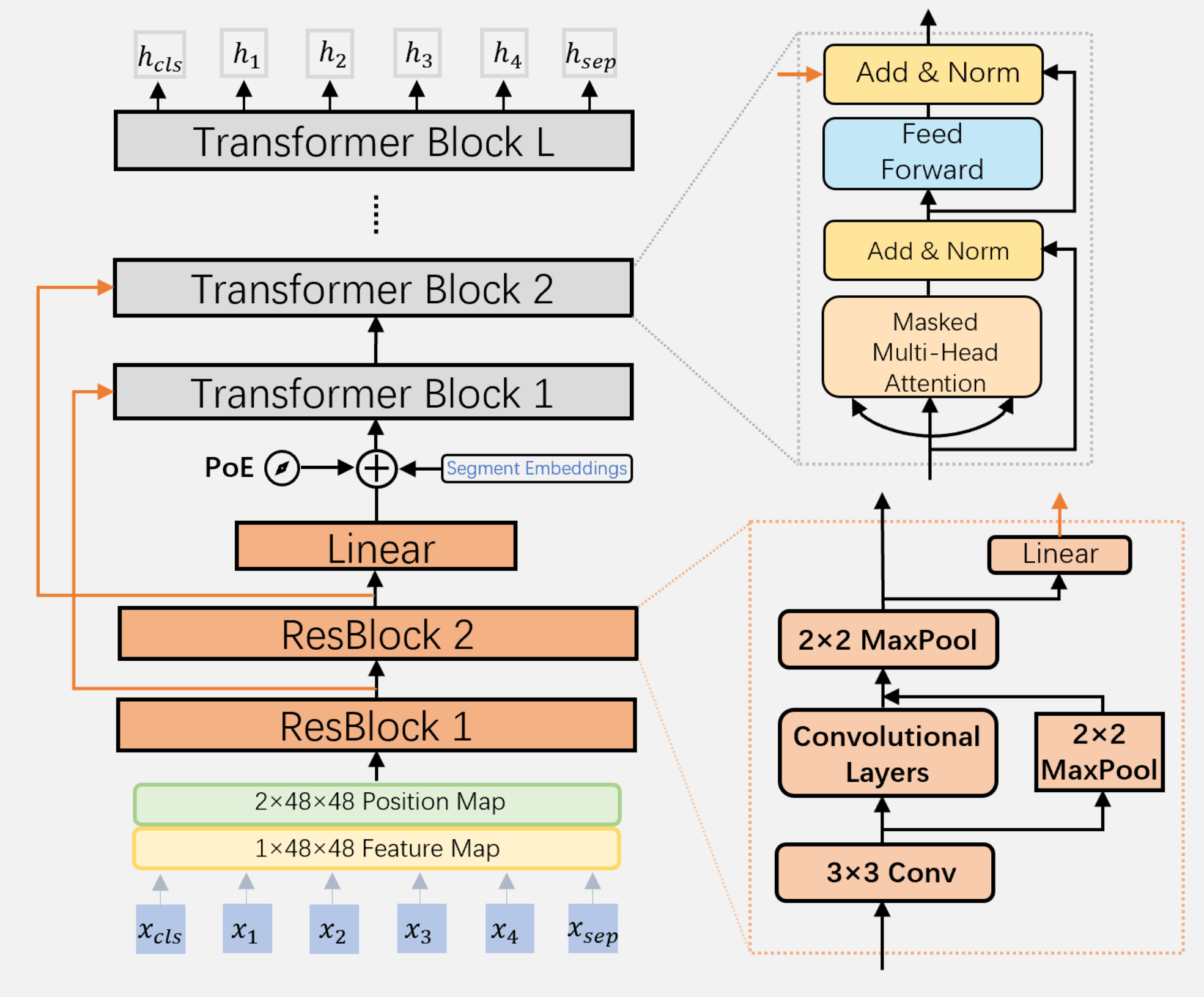}
    \caption{The overview of GlyphCRM. The HanGlyph module contains two identical residual convolutional blocks, yet with different parameters, namely 'ResBlock 1' and 'ResBlock 2'. 'PoE' is the position embeddings designed for the input sequence. The Transformer Block is used to capture the bidirectional contextual representation, including the Multi-Head Attention and Feed Forward Neural Network.}
    \label{fig:model}
\end{figure}

After obtaining the three-channel representation of each Chinese input character as shown in Figure \ref{fig:axis}, we adopt two residual convolutional blocks to extract its glyph feature, namely HanGlyph. As the orange box shown in Figure~\ref{fig:model}, for each residual convolutional block, the sequential input feature maps pass one convolutional layer with $3\times3$ kernel size, three-layer CNN and $2\times2$ max-pooling layer in turn. We take the second residual convolutional block for instance, which can be calculated by the equation as E.q.\ref{eq1}.
\begin{equation}
    \begin{array}{c}
    \mathbf{z}_{c}^{2} = \mathbf{ReLU}(\mathbf{w}^{2}\mathbf{z}^{1} + \mathbf{b}^{2})\vspace{1ex} \\
    \mathbf{z}_{r}^{2} = \mathbf{ReLU}( \mathbf{Max}_{2\times2}(\mathbf{z}_{c}^{2}) + \mathcal{F}(\mathbf{z}_{c}^{2}, \mathbf{W}_{r}^{2})) \vspace{1ex}\\
    \mathbf{z}^{2} = \mathbf{Max}_{2\times2}(\mathbf{z}_{r}^{2})
    
    \end{array}
    \label{eq1}
\end{equation} where $\mathbf{z}^{1}$ is the output of ResBlock 1 and $\mathbf{ReLU}$~\cite{relu} is the activation function widely used for deep convolutional neural networks (DCNN). $\mathbf{w}^{2}$ and $\mathbf{b}^{2}$ are the parameters for the first convolutional sub-layer where the padding width and stride length both are $1$ for each residual block. The output of the first sub-layer passes the 
three-layer CNN denoted as $\mathcal{F}(\mathbf{z}_{c}^{2}, \mathbf{W}_{r}^{2})$. 
Note that the core three-layer convolutional networks of ResBlock 1 have a large kernel window $9\times9$ to alleviate the issue of sparse image features, which is caused by the strokes of most Chinese characters occupying only a small number of pixels. Finally, one $2\times2$ max-pooling layer is used to obtain the final output for each residual block. 

Besides entering the next layer, we convert the final output of each residual block into one hidden state with the same dimensions as the upper bidirectional encoder layer through the linear layer. We denote the hidden state to $\mathbf{G}=(\mathbf{g_1}, \mathbf{g_2}, ..., \mathbf{g_n})$, which can be directly fed into the Transformer block as shown in Figure~\ref{fig:model}. ResBlock 2 is followed by one linear function to convert the output of each Chinese character into the glyph vector with fixed dimensions. It allows us to obtain the sequential glyph vectors of Chinese text, denoted as $\mathbf{R}=(\mathbf{r}_1, ..., \mathbf{r}_i, ..., \mathbf{r}_n)$, where $\mathbf{r}_i$ is the single Chinese character representation and $n$ is the length of input.

\subsection{Bidirectional Encoder Layer}

After obtaining the glyph vector of each Chinese character through the HanGlyph module, we use the bidirectional encoder layer based on Transformer\cite{attention} to obtain the context-sensitive representation of Chinese characters. 
Before $R$ is input into the superstructure, we sum the sequential glyph vectors, position embeddings, and segment embeddings to construct the more reasonable input; otherwise, characters in all input positions are regarded as equally important by the attention mechanism. Yet Chinese characters in different positions usually play different roles for understanding sentences.       

We adopt the popular Transformer structure as the backbone of the bidirectional encoder layer due to its capability to capture long-distance dependency information between words~\cite{devlin-etal-2019-bert, debert, T5}. Concretely, each Transformer block, shown in Figure~\ref{fig:model}, includes the multi-head attention mechanism and feed-forward neural networks. We take the $l$ th Transformer block as an instance and the computation process can be presented by the following equations.
\begin{equation}
\begin{array}{c}
\mathbf{h}_{l}^{M}=\mathbf{LayerNorm}(\mathbf{h}_{l-1} + \mathbf{MultiHeadAttention}(\mathbf{h}_{l-1}))\vspace{1ex} \\
\mathbf{h}_{l} = \mathbf{LayerNorm}(\mathbf{h}_{l}^{M} + \mathbf{FFN}(\mathbf{h}_{l}^{M}))  
\end{array}
\label{eq2}
\end{equation} where $\mathbf{h}_{l-1}$ is the output of the $l-1$ th Transformer block. $\mathbf{LayerNorm}$ \cite{layernorm} is one way to reduce the training time by performing normalization on the feature dimension of input. $\mathbf{FFN}$ is the feed-forward neural network, which is similar to Multi-Layer Perceptrons~\cite{perceptron}. $\mathbf{MultiHeadAttention}$ is the core structure of each Transformer blo-ck, which is used to update the contextual representation of characters by way of calculating the similarity with aspect to other characters. Specifically, the process can be denoted as E.q.\ref{eq3}.
\begin{equation}
\begin{array}{c}
    \mathbf{MultiHeadAttention} = \mathbf{Concat}(head_1, head_2, ..., head_h)\mathbf{W}^{h}\vspace{1ex} \\
    head_i = \mathbf{Attention}(\mathbf{Q}\mathbf{W}_{i}^{Q}, \mathbf{K}\mathbf{W}_{i}^{K}, \mathbf{V}\mathbf{W}_{i}^{V})
\end{array}    
\label{eq3}
\end{equation} where $h$ is the number of heads. $\mathbf{W}_{i}^{Q}$, $\mathbf{W}_{i}^{K}$ and $\mathbf{W}_{i}^{V}$ are the projection parameters, and their dimensions are $\mathbb{R}^{d_{model}\times d_{h}}$. $\mathbf{W}^{h}\in \mathbb{R}^{h d_{h} \times d_{model}}$, where $d_{model}$ is the hidden size. In our work, we set $d_h = d_{model}/h$ for each Transformer block. The dimension of each head is reduced, so the total computational cost is almost identical to the single head attention with full dimensionality. For the dot-product $\mathbf{Attention}$, the computational method is denoted as E.q.\ref{eq4}.
\begin{equation}
    \mathbf{Attention(\mathbf{Q},\mathbf{K},\mathbf{V}) = \mathbf{softmax}(\frac{\mathbf{Q}\mathbf{K}^{T}}{\sqrt{d_{k}}})\mathbf{V}}
    \label{eq4}
\end{equation}
In practice, the queries $\mathbf{Q}$, keys $\mathbf{K}$ and values $\mathbf{V}$ present the whole input sequence in each block. However, when updating the representation of any character, the query vector is itself, and the keys and values are the hidden states of the whole input sequence.

\begin{figure}
    \centering
    \includegraphics[width=0.35\textwidth]{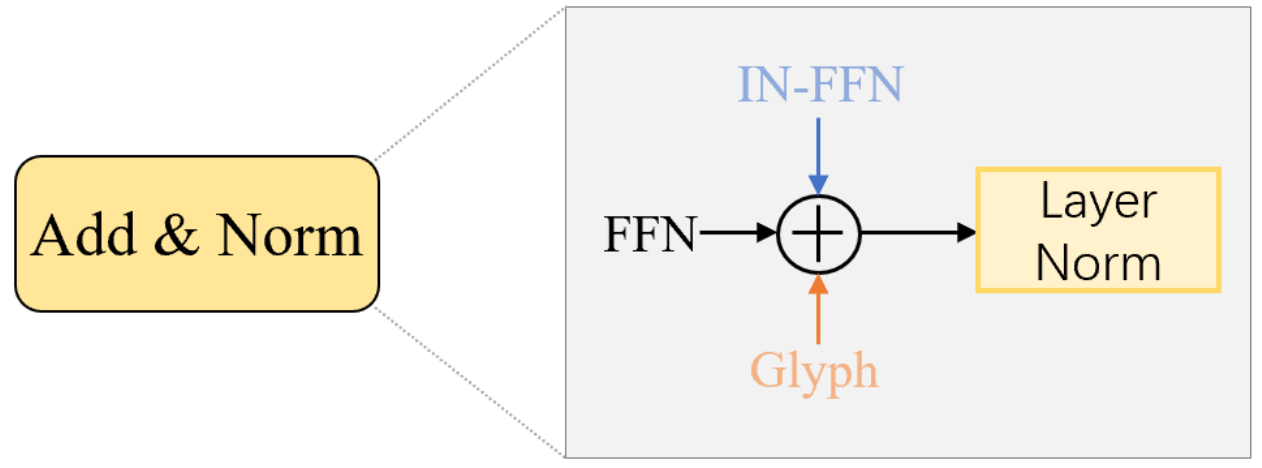}
    \caption{The second Add and Norm sub-module of the bottom two Transformer blocks. 'IN-FFN' represents the hidden state before being input into the feed-forward neural network. 'Glyph' represents the glyph vector obtained in the HanGlyph layer. 'FFN' represents the output of feed-forward neural network.}
    \label{fig:add}
\end{figure}

In order to make full use of the glyph features of Chinese characters, we incorporate the glyph representation obtained in each residual block of the HanGlyph module into the Transformer Block of the first two layers. As the gray box shown in Figure~\ref{fig:model}, the output of the FFN of the underlying two Transformer blocks is not only integrated with the output of the multi-head attention module, but also with the output of the residual blocks of HanGlyph. The detailed process is presented in Figure~\ref{fig:add}. Concretely, we take the simple addition of the three input vectors because an overly complex integration method will significantly increase the computation cost of GlyphCRM, and this method is also proved effective~\cite{resnet}.

From the overall architecture of GlyphCRM, the underlying four-layer networks are a symmetrical structure, which can have a stable information interaction way to capture and exploit glyph features. To summarize, GlyphCRM contains the glyph features of Chinese characters extracted from images and the contextual information of input text, which can be regarded as a pre-trained multi-modal model but is different from general cross-modality models such as VisualBERT~\cite{visualbert}, LXMERT~\cite{tan-bansal-2019-lxmert} and ERNIE-ViL~\cite{ernie_vil}. Without using the ID-based word embedding method, GlyphCRM will not be restricted by the unseen (out of fixed vocab) Chinese characters when fine-tuned on specific downstream tasks which contain unseen characters. Furthermore, the glyph features of Chinese characters can be used to infer the meaning of previously unseen or sparse Chinese characters by glyph similarity, compared to directly converting them into word vector according to word/character ID having no such inference information. 

\subsection{Training}
In this section, we introduce the detailed two-stage training process of our model, containing pre-training and fine-tuning.

\subsubsection{Pre-training} 
To avoid overfitting, models with huge parameters usually need large-scale corpus to train, but manual labeling will cost copious resources. Yet like BERT, the large neural language models are usually pre-trained on large-scale corpora with an unsupervised method to enable them have a detailed understanding of texts on specific languages, i.e., pre-trained language models have good initial parameters when used for specific tasks again. Hence, training on large-scale corpora with unsupervised methods is a relatively efficient approach for training large-param-eter models due to the unnecessary to spend enormous manpower and material resources, thereby achieving a remarkable success in natural language processing.      

\begin{figure}[t]
    \centering
    \includegraphics[width=0.43\textwidth]{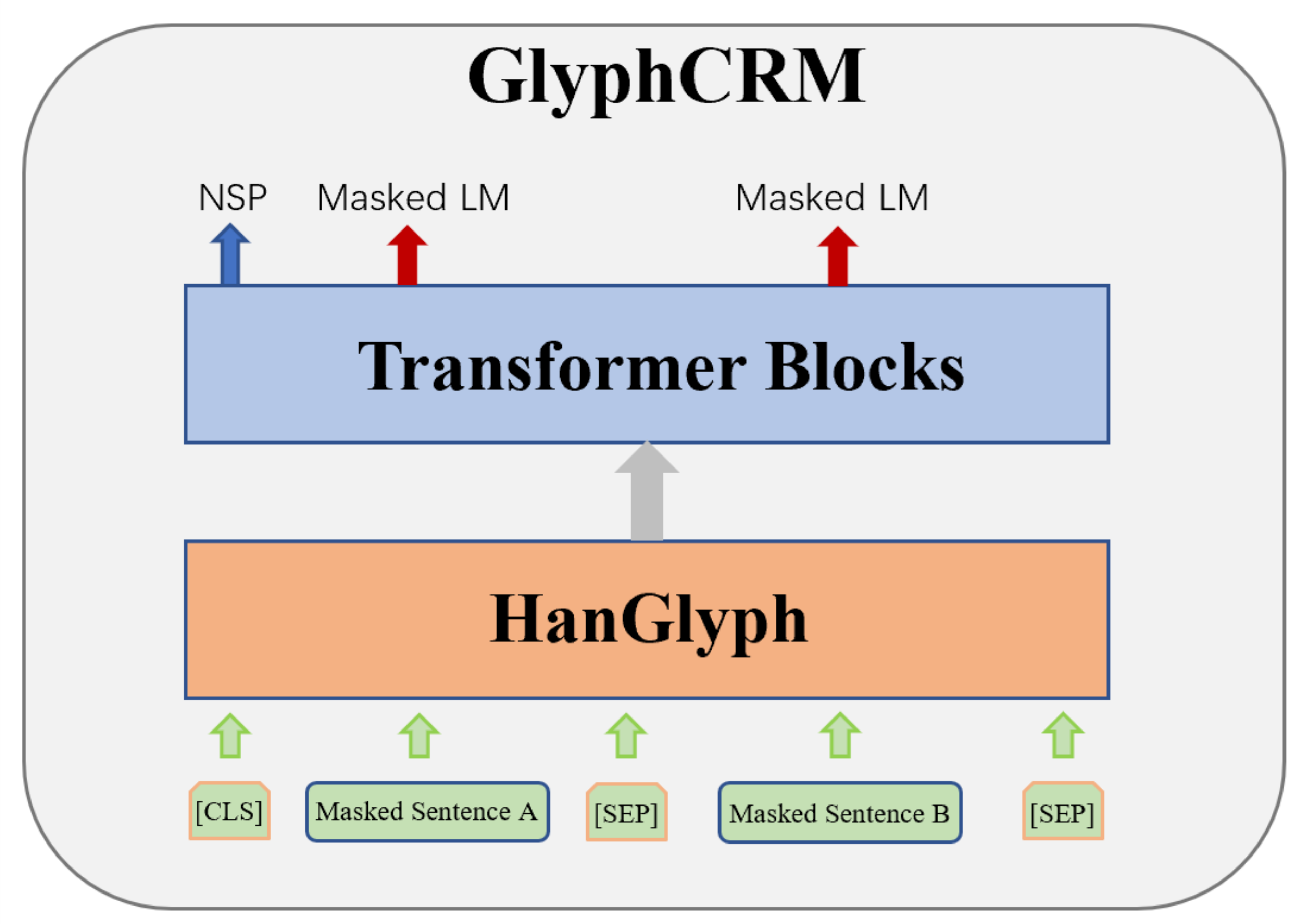}
    \caption{The overview of unsupervised next sentence prediction~(NSP) and masked language model~(Masked LM). The start symbol $\rm[CLS]$ is used for predicting the relationship between sentence A and B. $\rm[SEP]$ is used to divide sentences.}
    \label{fig:nsp}
\end{figure}

In this paper, we first pre-train our proposed model GlyphCRM with two unsupervised tasks, identical to those in BERT~\cite{devlin-etal-2019-bert}. The first one is the Masked LM task where we randomly choose 15\% of all Chinese characters for each input text to be replaced with one special token $\rm[MASK]$ 80\% of the time, a random character 10\% of the time, and the unchanged character 10\% of the time. It exploits other Chinese characters to predict the corrupted characters. For predicting the masked characters, we count the number of Chinese characters in the pre-training corpora and construct the corresponding vocabulary used for classification. 
We select the universally used Chinese characters which can be presented by Song typeface, monofont, or boldface. The sparse characters are replaced by another special token $\rm[UNK]$. The final vocab size is about $18,612$, less than that of BERT.
Specifically, for any input sequence $\mathbf{X}$, we first construct the corrupted input version $\hat{\mathbf{x}}$ by the above randomly masked method. We define the masked characters as $\overline{\mathbf{x}}$, and the training object of masked prediction can be presented as the following:
\begin{equation}
\mathcal{L}_{mp} = \max _{\theta} \log p_{\theta}(\overline{\mathbf{x}} \mid \hat{\mathbf{x}}) \approx \sum_{t=1}^{n} m_{t} \log \left(\mathbf{H}_{\theta}(\hat{\mathbf{x}})_{t} \mathbf{e}\left(x_{t}\right)\right)
\label{eq5}
\end{equation}
where $m_{t} = 1$ represents $x_{t}$ is masked, and $\mathbf{H}_{\theta}$ is the top hidden state of our model. Thus, the final hidden states of the input text can be denoted as $\mathbf{H}_{\theta}(\mathbf{X}) = (\mathbf{H}_{\theta}(\mathbf{x})_{1}, \mathbf{H}_{\theta}(\mathbf{x})_{2}, ..., \mathbf{H}_{\theta}(\mathbf{x})_{n})$. $\mathbf{e(x)}$ indicates the final projection matrix that maps the final hidden state of masked characters to the vocabulary size.

Besides, we adopt the next sentence prediction~(NSP) pre-training task to impel our model to understand the relationship between sentences, which is instrumental for being fine-tuned on some downstream tasks such as Question Answering~\cite{questiona2, wang-etal-2017-questiona} and Sentence Matching~\cite{sentence_m, NIPS2014_hu}. Concretely, when the training example is the composition of sentence A and B, we formulate that 50\% of the time B is the general next sentence that follows A~(regarded as the IsNext), 50\% of the time B is from the other training data~(regarded as the NotNext). While training, the top hidden state of the start token $\rm[CLS]$ is used to predict the relationship between two sentences as the blue arrow shown in Figure~\ref{fig:nsp}.

\subsubsection{Fine-tuning} Compared to pre-training, fine-tuning expends relatively fewer resources. In this paper, we fine-tune our proposed model on general natural language processing tasks, including single sentence classification, text classification, sequence labeling and sentence matching. For single sentence classification, sentence mat-ching, and multi-sentence text classification tasks, we feed the final output of the start token $\rm[CLS]$ to one task-specific output layer to predict the correct label. For sequence labeling, the final output of each Chinese character is fed into an output layer for classification. Notably, the self-attention mechanism in the Transform block of GlyphCRM ensures the almost seamless connection of the two stages of pre-training and fine-tuning, making the application of our model direct and effective in specific tasks. The detailed analyses and comparison with BERT will be presented in the following experiment section.

\section{Experiment}
In this section, we first introduce the detailed experimental settings and the pre-training performance of our model. Secondly, we in-detail analyze the performance of GlyphCRM on 9 Chinese natural language understanding~(NLU) datasets.   

\begin{figure}[t]
    \centering
    \includegraphics[width=0.43\textwidth, height=0.30\textwidth]{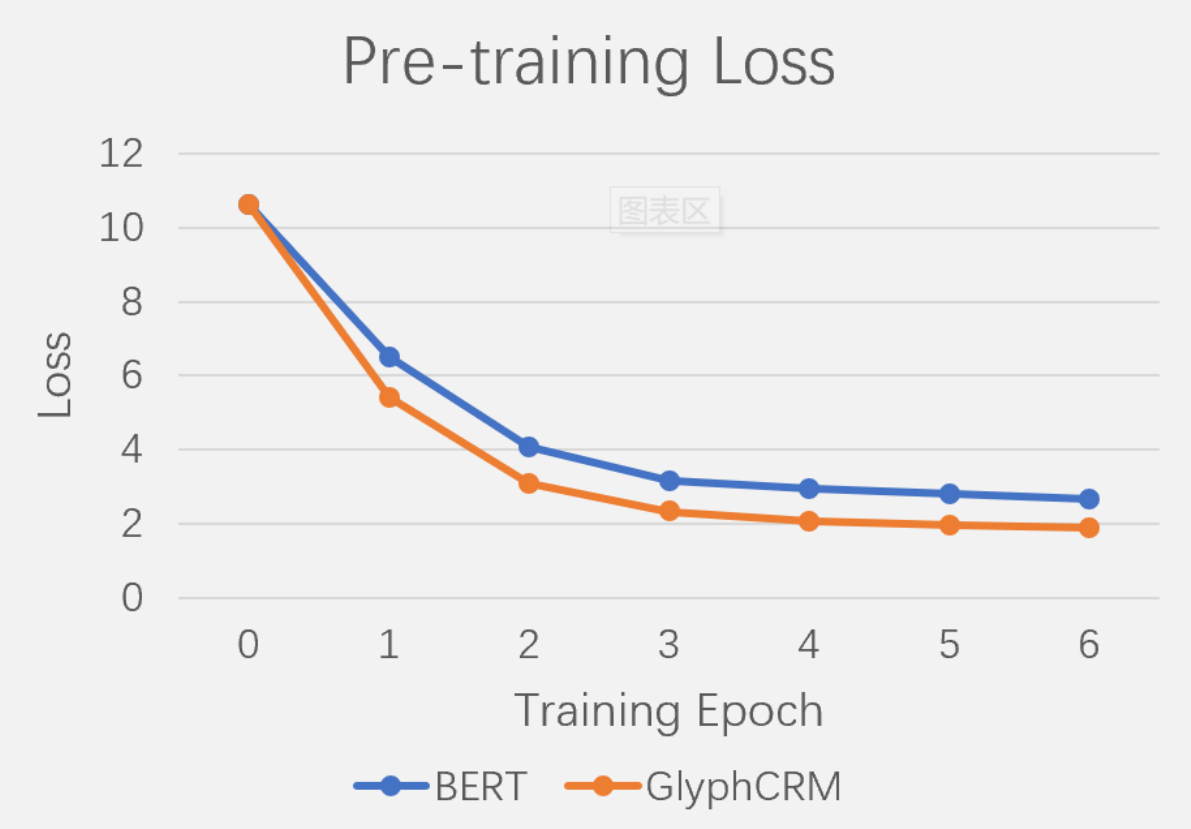}
    \caption{The pre-training loss curves of BERT and GlyphCRM in the first 6 epochs. Each point in the curve represents the average loss of current epoch. Epoch '0' represents the starting point of training. }
    \label{fig:loss}
\end{figure}

\subsection{Experimental settings}

\subsubsection{Model Architecture} The architecture of GlyphCRM and the baseline model BERT-Base we adopt both have 12 Transformer blocks. Each layer of them has 12 attention heads, and the size of hidden states is $768$. The total parameters of BERT-Base are \textbf{110} million, yet GlyphCRM only has \textbf{95} million parameters. In the case of the same number of Transformer blocks, the proposed model has fewer parameters. We separately pre-train BERT and GlyphCRM for $\textbf{15}$ and $\textbf{6}$ epochs\footnote{ Due to the limited computational resources and from the downward trend of the loss curve of GlyphCRM, we adopt it with being pre-trained 6 epochs for fine-tuning.} on the same processed $\textbf{3}$ million pre-trained Chinese corpora. From the downward trend of the two pre-training losses shown in Figure \ref{fig:loss}, we can observe that the representation model based entirely on glyphs of Chinese characters has a stronger learning ability than BERT.

\subsubsection{Datasets for Evaluation} We first compare GlyphCRM and BERT on the following Chinese NLU datasets. 

\noindent\textbf{ChnSentiCorp:} ChnSentiCorp~\cite{chnSentiCorp} is the Chinese sentiment analysis dataset, including three-domain documents: education, movie, and house. Each domain contains two classification labels: positive and negative. We divide the data in each domain into training, validation, and test sets at the ratio of 0.8:0.1:0.1.

\noindent\textbf{Hotel Review Sentiment Analysis:} Hotel Review Sentiment An-alysis dataset\footnote{https://pan.baidu.com/s/1Y4vPSSH4ukPfO4ONUg2lSg} is collected from the Ctrip website, including 10k texts and two positive and negative classification labels. It is an unbalanced sentiment classification dataset, having 7k positive samples. We divide the data of each label into training, validation, and test sets at the ratio of 0.8:0.1:0.1. 

\noindent\textbf{Chinese Natural Language Inference:} The Chinese Natural Language Inference~(CNLI) is from the public evaluation tasks of the Seventeenth China National Conference on Computational Linguistics\footnote{http://www.cips-cl.org/static/CCL2018/call-evaluation.html} (CCL2018). We split the whole dataset into $90,000$, $10,000$, $10,000$ as the train, validation and test set. Each example contains two sentences, where the relationship between them is entailment, contradiction, or neural.

\noindent\textbf{THUCNews:} THUCNews~\cite{thunews} is a long document classification dataset. After processing the long document dataset, we retain 65k documents. We randomly select 10k samples as the test set, 5k samples as the validation set, and 50k samples as the training set. 

\noindent\textbf{TouTiaoCNews:} TouTiaoCNews~\footnote{https://github.com/aceimnorstuvwxz/toutiao-text-classfication-dataset} is the short news text classification dataset. It contains $382,688$ examples, and all documents are divided into 15 categories according to their content. We randomly divide the total data with a ratio of 0.8:0.1:0.1, separately as the train, validation, and test set. 

\noindent\textbf{MSRA-NER:} MSRA-NER~\cite{msra, nermsra} dataset is the sequence labeling task proposed by Microsoft Research Asia in 2006. It contains 7 tagging labels: O, B-PER, I-PER, B-ORG, I-ORG, B-LOC, and I-LOC. Hence, it can be regarded as the 7 classification task.

\noindent\textbf{People-NER:} The data of People-NER comes from the article of the People's Daily in 2014~\footnote{https://pan.baidu.com/s/1LDwQjoj7qc-HT9qwhJ3rcA}. It contains approximately 28k data, and has the same 7 tagging labels as MSRA-NER. We split it into train, validation, and test set, separately with $20,864$, $2,318$, and $4,636$ samples.

\noindent\textbf{Unknown character Statistics: } We count the distribution of unknown characters in the above 7 datasets according to the vocab of BERT. As the results shown in Table~\ref{tab:exp5}, despite the vocab of BERT has $21,128$ tokens, there are still many unknown characters in the long Chinese texts. Even for fine-grained classification tasks such as MSRA-NER and People-NER, there are usually many unknown characters in the dataset, so it is meaningful and valuable to solve the out-of-vocabulary problem.

\begin{table}[t]
\renewcommand\arraystretch{1.1}
\small
\centering
 \caption{The detailed statistics of unknown characters (chars) in different datasets.}
\label{tab:exp5}
\begin{tabular}{lccc}
\toprule[0.9pt]
\multicolumn{1}{c}{\textbf{Dataset}} &  \multicolumn{1}{c}{\textbf{Unknown Chars}} & \multicolumn{1}{c}{\textbf{Total Chars}} &
\multicolumn{1}{c}{\textbf{Ratio~(\%)}}\\\midrule[0.7pt]
\multicolumn{4}{c}{\textbf{Validation}} \\\hline
\# ChnSentidev & $798$ & $123,568$ & $0.65$\\
\# CNLI & $722$ & $288,534$ & $0.25$\\
\# HRSA & $358$ & $123,690$ & $0.29$\\
\# THUCNews & $34,604$ & $2,004,771$ & $1.72$\\
\# TouTiaoCNews & $26,750$ & $1,384,652$ & $\textbf{1.93}$\\
\# MSRA-NER & $1,508$ & $142,095$ & $1.06$\\
\# People-NER & $1,061$ & $107,894$ & $0.98$\\
\midrule[0.7pt]
\multicolumn{4}{c}{\textbf{Test}} \\\hline
\# ChnSentidev & $790$ & $120,412$ & $0.66$\\
\# CNLI & $754$ & $285,167$ & $0.26$\\
\# HRSA & $322$ & $118,694$ & $0.27$\\
\# THUCNews & $69,686$ & $4,247,903$ & $1.64$\\
\# TouTiaoCNews & $26,805$ & $1,380,889$ & $\textbf{1.94}$\\
\# MSRA-NER & $1,468$ & $169,751$ & $0.86$\\
\# People-NER & $2,137$ & $215,530$ & $0.99$\\
\bottomrule[0.9pt]
\end{tabular}
\end{table}

\subsubsection{Experimental Details} We use 2 Tesla V100 GPUs to pre-train BERT and GlyphCRM on the processed 3 million Chinese texts with Adam~\cite{kingma2014method} optimizer with the initial learning rate to 0.0001, $\beta_{1} = 0.9$, $\beta_{2} = 0.999$.
The pre-training Chinese texts comes from the Chinese Wikipedia\footnote{https://dumps.wikimedia.org/zhwiki/}, and the preprocessing way is identical to BERT.
We set the weight decay to 0.01 and set the linear decay of learning rate, and the learning rate warmup over the first 10k steps.
We set the batch size and maximum input length to 256 and 512. The pre-training loss is the sum of the mean Masked LM likelihood and the mean NSP likelihood. Furthermore, we fine-tune GlyphCRM and BERT on the specific tasks with the same hyperparameter settings.
We set up different learning rates for different tasks, which are presented in the following experiment analyses. 

\subsection{Results and Analysis}
In this section, we will represent the comparison results between the previous state-of-the-art pre-trained model BERT and our designed model GlyphCRM. Note that we use the validation set to select the model with the best performance.


\subsubsection{Single Sentence Classification} ChnSentiCorp and Hotel Review Sentiment Analysis are coarse-grained sentiment classification datasets, where models are trained to perform sentence-level binary classification task. The evaluation metrics for all datasets are the prediction accuracy. For classification, we enable the last hidden state of $\rm[CLS]$ pass a fully connected linear layer followed by a softmax activation function to classify the input sentence. The experimental results are shown in Table~\ref{tab:exp1}.


The above experimental results show that whether it is from the validation set or the test set, the classification accuracy of our model significantly exceeds BERT on the two sentiment classification dataset, especially on the test set. For instance, $93.08~\rm{vs}~91.25$ on the test set of ChnSentiCorp and  $92.70~\rm{vs}~91.40$ for Hotel Review Sentiment Analysis. It demonstrates that GlyphCRM has a stronger ability to understand the semantics of whole sentences compared to BERT. Moreover, the experimental results in Table~\ref{tab:exp1} and loss curves in Figure~\ref{fig:loss} indicate that our model can learn the contextual information of the overall sentence better and faster. It may be attributed to the fact that the glyph features extracted by the HanGlyph module can sufficiently express the structure of Chinese characters, which has the easily distinguishable features.
In addition, GlyphCRM could distinguish the contextual semantics of positive and negative words according to the glyph features of them.  
Overall, the above analyses and the experimental results of comparison models further indicate that glyph features of Chinese characters are practical when used in the sentiment analysis task.



\begin{table}[t]
\renewcommand\arraystretch{1.1}
    \centering
     \caption{Automatic evaluation results on the validation and test sets of ChnSentiCorp and Hotel Review Sentiment Analysis. The leftmost column represents the model and their pre-training degree. LR represents the learning rate of a specific task~(Unless otherwise stated, the instruction LR in the following table are the same). $15e$ and $6e$ separately represent the pre-training epochs of BERT and GlyphCRM.}
    \label{tab:exp1}
    \begin{tabular}{lccc}
    \toprule[0.9pt]
    \multicolumn{1}{c}{\textbf{Model}} & \multicolumn{1}{c}{\textbf{LR}} & \multicolumn{1}{c}{\textbf{Dev ACC(\%)}} & \multicolumn{1}{c}{\textbf{Test ACC(\%)}} \\
    \midrule[0.7pt]
    \multicolumn{4}{c}{\textbf{ChnSentiCorp}}\\ \hline
    \# BERT$_{15e}$ & $2e-5$ & $92.66$ & $91.25$ \\
    \# GlyphCRMT$_{6e}$ & $2e-5$ & $\textbf{93.17}$ & $\textbf{93.08}$\\\midrule[0.7pt]
    \multicolumn{4}{c}{\textbf{Hotel Review Sentiment Analysis}}\\ \hline
    \# BERT$_{15e}$ & $2e-5$ & $92.88$ & $91.40$ \\
    \# GlyphCRM$_{6e}$ & $2e-5$ & $\textbf{93.80}$ & $\textbf{92.70}$\\
    \bottomrule[0.9pt]
    \end{tabular}
\end{table}


\subsubsection{Chinese Text Classification} After evaluating the performan-ce of models on the single sentence classification task, we tested them on two multi-sentence document classification datasets THU-CNews and TouTiaoCNews to evaluate their ability to understand Chinese long documents. Formally, we still take a fully connected linear layer followed by a softmax activation function to map the classification labels. The experimental results on validation and test sets are shown in Table~\ref{tab:exp2}. From the experimental results, we observe that the performance of GlyphCRM is still better than BERT, yet the gap between two models is smaller compared to being evaluated on the single sentence classification task. Specifically, the performance of our model separately exceeds BERT by about $0.5\%$ and $0.75\%$ on the validation and test sets, yet GlyphCRM outperforms BERT by about $1.5\%$ gains on two test sets as shown in Table~\ref{tab:exp1}. 
It indicates that the performance of our model based entirely on glyphs is comparable to and even surpasses BERT when dealing with long Chinese documents. Furthermore, whether from short text (at least one sentence) or long text, the pre-trained representation model based entirely on Chinese glyphs is simple and effective for Chinese multi-label classification tasks. 


\begin{table}[t]
\renewcommand\arraystretch{1.1}
    \centering
     \caption{Automatic evaluation results on the validation and test sets of THUCNews and TouTiaoCNews. }
    \label{tab:exp2}
    \begin{tabular}{lccc}
    \toprule[0.9pt]
    \multicolumn{1}{c}{\textbf{Model}} & \multicolumn{1}{c}{\textbf{LR}} & \multicolumn{1}{c}{\textbf{Dev ACC(\%)}} & \multicolumn{1}{c}{\textbf{Test ACC(\%)}} \\\midrule[0.7pt]
    \multicolumn{4}{c}{\textbf{THUCNews}}\\ \hline
    \# BERT$_{15e}$ & $3e-5$ & $96.28$ & $95.41$ \\
    \# GlyphCRM$_{6e}$ & $3e-5$ & $\textbf{96.60}$ & $\textbf{96.32}$\\ \midrule[0.7pt]
    \multicolumn{4}{c}{\textbf{TouTiaoCNews}}\\ \hline
    \# BERT$_{15e}$ & $3e-5$ & $88.85$ & $88.84$ \\
    \# GlyphCRM$_{6e}$ & $3e-5$ & $\textbf{89.60}$ & $\textbf{89.45}$\\ \bottomrule[0.9pt]
    \end{tabular}
\end{table}

\subsubsection{Chinese Natural Language Inference} Natural language inference is mainly to determine whether there is an inference implication relationship between two sentences. It further investigates the model's ability to understand sentence semantics and has high transferability in real-word applications. We select a classic Chinese language inference evaluation dataset presented in CCL2018. We conduct the experiment with an initial learning rate $3e-5$, and the experimental results are shown in Table \ref{tab:exp3}.
\begin{table}[H]
\renewcommand\arraystretch{1.1}
    \centering
     \caption{Automatic evaluation results on the validation and test sets of Chinese natural language inference~(CNLI).}
    \label{tab:exp3}
    \begin{tabular}{lccc}
    \toprule[0.9pt]
    \multicolumn{1}{c}{\textbf{Model}} & \multicolumn{1}{c}{\textbf{LR}} & \multicolumn{1}{c}{\textbf{Dev ACC(\%)}} & \multicolumn{1}{c}{\textbf{Test ACC(\%)}} \\\midrule[0.7pt]
    \# BERT$_{15e}$ & $3e-5$ & $71.04$ & $69.74$ \\
    \# GlyphCRM$_{6e}$ & $3e-5$ & $\textbf{72.28}$ & $\textbf{71.80}$\\
    \bottomrule[0.9pt]
    \end{tabular}
\end{table}
\begin{CJK}{UTF8}{gbsn}
\begin{table}[H]
\renewcommand\arraystretch{1.15}
\small
\centering
\caption{Two instances of the test set of CNLI.}
\label{tab:exp6}
\begin{tabular}{|p{7.5cm}|}
\hline
\multicolumn{1}{|c|}{\textbf{CNLI (test)}} \\
\hline
$\bullet$~\textbf{Sentence A:} 一位黑发\textbf{女子}在\textbf{舞台}上用相机拍下了一张\textbf{乐队}的照片。(A black-haired \textbf{woman} took a photo of the \textbf{band} with a camera on the \textbf{stage}.)\\
$\bullet$~\textbf{Sentence B:} 一位\textbf{女士}正在一场\textbf{音乐会}上。(A \textbf{lady} is at a \textbf{concert}.)\\ 
$\bullet$~\textbf{Label:} \textcolor{blue}{Entailment}\\
$\bullet$~\textbf{BERT:} \textcolor{red}{Neural}\\
$\bullet$~\textbf{GlyphCRM:} Entailment\\
\hline
$\bullet$~\textbf{Sentence A:} 一位有才华的男性艺术家正在外面的\textbf{伞}下\textbf{画漫画}。(A talented male artist is \textbf{drawing a comic} outside under an \textbf{umbrella}.)\\
$\bullet$~\textbf{Sentence B:} 这位艺术家在\textbf{购物中心}内的\textbf{售货亭}\textbf{工作}。(This artist works at the \textbf{kiosk} in the \textbf{shopping center}.)\\ 
$\bullet$~\textbf{Label:} \textcolor{orange}{Contradiction}\\
$\bullet$~\textbf{BERT:} \textcolor{red}{Neural}\\
$\bullet$~\textbf{GlyphCRM:} Contradiction\\
\hline
\end{tabular}
\end{table}
\end{CJK}
Table~\ref{tab:exp3} shows that our model separately surpasses the benchmark model BERT by about 1.2\%, 2\% on the validation and test set, demonstrating that full glyphs of Chinese characters are expressive enough to be used for their representations.
The excellent performance may be attributed to two facts: 1) GlyphCRM can further capture sentence-level semantic according to the glyph features of consecutive Chinese characters. 2) Compared to directly converting the character into a vector, each Chinese character representation incorporating its glyph is more distinguishable in the semantic space.

\begin{CJK}{UTF8}{gbsn}
Moreover, we select two examples from the test set to verify the performance of models, presented in Table~\ref{tab:exp6}. For the first instance, we observe that the first sentence can deduce the second sentence according to the keywords of the first sentence: '女子'~(women), '舞台'(stage), '乐队'~(band) and keywords of the second sentence: '女士'~(lady), '音乐会'~(concert). For the second instance, the semantics of keywords in the sentence B such as '购物中心'~(shopping center), '售货亭'~(kiosk) and '工作'~(working), are different from '伞'~(umbrella) and '画漫画'~(drawing comic) in sentence A. Judging from the inference results of models, BERT usually cannot infer the relationship between two sentences in many cases, thus determining that the two sentences are neural. However, our model incorporating glyphs has a great advantage in inferring the semantic relevance between sentences. 
\end{CJK}

\begin{table}[t]
\renewcommand\arraystretch{1.1}
    \centering
     \caption{Automatic evaluation results on the validation and test sets of Chinese named entity recognition tasks, including MSRA-NER and People-NER. F1, P, and R represent F1-score, Precision, and Recall.}
    \label{tab:exp4}
    \begin{tabular}{lcccc}
    \toprule[0.9pt]
    \multicolumn{1}{c}{\textbf{Model}} & \multicolumn{1}{c}{\textbf{LR}} & \multicolumn{1}{c}{\textbf{F1(\%)}} & \multicolumn{1}{c}{\textbf{P(\%)}}&
    \multicolumn{1}{c}{\textbf{R(\%)}}\\\midrule[0.7pt]
    \multicolumn{5}{c}{\textbf{MSRA-NER~(validation)}}\\\hline
    \# BERT$_{15e}$ & $3e-5$ & $83.55$ & $81.69$& $85.48$\\
    \# GlyphCRM$_{6e}$ & $3e-5$ & $\textbf{90.15}$ & $\textbf{89.32}$& $\textbf{91.00}$\\\midrule[0.7pt]
    \multicolumn{5}{c}{\textbf{MSRA-NER~(test)}}\\ \hline
    \# BERT$_{15e}$ & $3e-5$ & $77.78$ & $75.84$ &$79.81$\\
    \# GlyphCRM$_{6e}$ & $3e-5$ & $\textbf{86.04}$ & $\textbf{85.53}$ &$\textbf{86.57}$\\ \midrule[0.7pt]
    \multicolumn{5}{c}{\textbf{People-NER~(validation)}}\\\hline
    \# BERT$_{15e}$ & $3e-5$ & $80.40$ & $78.61$& $82.27$\\
    \# GlyphCRM$_{6e}$ & $3e-5$ & $\textbf{85.62}$ & $\textbf{85.34}$& $\textbf{85.90}$\\\midrule[0.7pt]
    \multicolumn{5}{c}{\textbf{People-NER~(test)}}\\\hline
    \# BERT$_{15e}$ & $3e-5$ & $79.20$ & $76.91$& $81.64$\\
    \# GlyphCRM$_{6e}$ & $3e-5$ & $\textbf{84.05}$ & $\textbf{82.80}$& $\textbf{85.35}$\\
    \bottomrule[0.9pt]
    \end{tabular}
\end{table}

\subsubsection{Sequence Labeling} Compared to text classification tasks that focus on the overall understanding of input content, sequence labeling, e.g., word segmentation, part-of-speech tagging, named entity recognition, and relationship extraction, is more fine-grained classification tasks. It can be used to evaluate the ability of Chinese pre-trained representation models to represent Chinese characters, which is directly related to the fine-grained classification accuracy. Different from previous methods of adding CRF~\cite{laffertyCrf} network based on the pre-trained representation model, we just add a fully connected layer to the final output hidden states of models. So the whole architecture mainly relies on representation models. Notably, the final fully connected layer is exactly identical for GlyphCRM and BERT in order to compare them fairly. The performance of the whole network can directly evaluate the ability of the Chinese pre-trained representation model to express Chinese characters.
\begin{CJK}{UTF8}{gbsn}
\begin{table}[t]
\renewcommand\arraystretch{1.15}
\small
\centering
\caption{Two typical recognition instances of MSRA-NER and People-NER. 'O' represents that the character has no specific label as a category. 'B-LOC' and 'I-LOC' separately represent the start and inside of location words. Red-colored words are the error predictions. PER represents the name of Chinese people.}
\label{tab:exp7}
\begin{tabular}{|p{7.5cm}|}
\hline
\multicolumn{1}{|c|}{\textbf{MSRA-NER~(test)}} \\
\hline
$\bullet$~\textbf{Sentence}\\
这 是 一 座 典 型 的 加 泰 罗 尼 亚 民 居 ， 房 屋 的 建 筑 和 装 饰 风 格 与 周 围 民 居 没 有 什 么 区 别 。(This is a typical \textbf{Catalan house}, and the building and decoration style of the house is not different from the surrounding houses.)\\
$\bullet$~\textbf{True Label}\\
加 泰 罗 尼 亚：B-LOC I-LOC I-LOC I-LOC I-LOC\\ 
Others: O\\
$\bullet$~\textbf{BERT}\\
加 泰 罗 尼 亚 民 居: B-LOC I-ORG I-ORG I-ORG I-ORG \textcolor{red}{I-ORG I-ORG} \\
Others: O\\
$\bullet$~\textbf{GlyphCRM}\\
加 泰 罗 尼 亚：B-LOC I-LOC I-LOC I-LOC I-LOC\\ 
Others: O\\
\hline
\multicolumn{1}{|c|}{\textbf{People-NER~(test)}} \\
\hline
$\bullet$~\textbf{Sentence} \\
崔 剑 平 连 衣 服 也 没 顾 得 脱 ， 便 飞 身 跃 入 水 中 ， 奋 力 向 溺 水 者 游 去 。(\textbf{Cui, Jianping} didn't even take off his clothes, so he flew into the water and swam to the drowning person.)\\
$\bullet$~\textbf{True Label} \\
崔 剑 平: B-PER I-PER I-PER\\
Others: O\\
$\bullet$~\textbf{BERT}\\
崔 剑 平: B-PER I-PER \textcolor{red}{O}\\
Others: O\\
$\bullet$~\textbf{GlyphCRM}\\
崔 剑 平: B-PER I-PER I-PER\\
Others: O\\
\hline
\end{tabular}
\end{table}
\end{CJK}
We adopt the frequently used Chinese NER datasets MSRA-NER and People-NER to evaluate the fine-grained representation capability of GlyphCRM and BERT. The experimental results are shown in Table~\ref{tab:exp4}. Firstly, compared to the experimental results on Chinese text classification tasks, the performance of GlyphCRM and BERT has a greater gap in NER, e.g., F1-score: $90.15$ $\rm{vs}~83.55$ on MSRA validation, $86.04~\rm{vs}~77.78$ on MSRA test. Precision: $85.53$ $\rm{vs}$~$75.84$ on MSRA test, $82.80~\rm{vs}~76.91$ on People-NER test. Recall: $91.00~\rm{vs}~85.48$ on MSRA test, $85.35~\rm{vs}~81.64$ on People-NER test. It indicates that our model has a strong ability to understand the overall semantics of Chinese texts and excellent representation for the fine-grained (a single Chinese character) character semantics. Secondly, Table~\ref{tab:exp4} shows that the performance of our model is significantly higher than BERT in Precision and Recall, and the performance on the two indicators is not much different. It indicates that GlyphCRM has high overall recognition accuracy on sequence labeling tasks.

\begin{CJK}{UTF8}{gbsn}
Specifically, we select two typical examples from the test sets, presented in Table~\ref{tab:exp7}. 
BERT usually has inferior accuracy when predicting the ending position of location phrases. Yet, the glyph-based method can alleviate this problem by effectively learning the difference between phrases, e.g., the meaning conveyed by '民居'~(residence) and '加泰罗尼亚'~(Catalonia) and glyphs of them are different. The proposed model can predict it accurately yet BERT
give the wrong labels. 
Hence, the inferior performance of BERT on the People-NER containing many Chinese names is probably caused by inaccurately predicting Chinese names of triples or doubles as the second instance shown in Table \ref{tab:exp7}. Overall, GlyphCRM can learn the characteristics of Chinese phrases well according to the patterns of consecutive characters' glyphs on specific fine-tuning tasks.
\end{CJK}
\section{Transferability Assessment on Specialized Fields and Low-resource Tasks}

\textbf{Medical Field:} The transferability evaluation of pre-trained models has been attracting attention of many researchers because the high transferability means that models can adapt to a wide range of application scenarios. In this section, we first evaluate the two pre-trained models on the Chinese sentence semantic matching dataset in the medical field~(CMSSM), related to COVID-19. It is provided by More Health technology company\footnote{https://tianchi.aliyun.com/dataset/dataDetail?dataId=76751}. CMSSM is a fine-grained semantic matching task that mainly involves 10 diseases such as pneumonia, mycoplasma pneumonia, bronchitis, and so on. The length of each sentence is less than 20 words. Its training, validation and test set include $8,747$, $2,002$ and $7,032$ samples, respectively. 

\begin{CJK}{UTF8}{gbsn}
Table~\ref{tab:exp8} shows that the performance of GlyphCRM exceeds BERT by about 1\% both on the validation and test sets. It indicates that the performance of the Chinese pre-trained representation method based entirely on glyphs is excellent in the medical field. Generally speaking, the model pre-trained on large-scale open-domain corpora always performs poorly when fine-tuned in the specialized field~\cite{dont_stop_pretrain} due to exiting many professional terms. However, our pre-trained model can enlighten us to improve the model's transferability on specialized fields by introducing glyph vectors. After the retrospective analyses of CMSSM and the above 7 tasks, we observe that Chinese professional terms are usually composed of multiple Chinese pictographic characters, and their glyphs usually have certain links with the things to be described. For instance, '肺炎'~(Pneumonia) is usually caused by lung infections, and its symptom is fever. The glyph of '肺'~(lungs) is similar to the human organ lung, and '炎'~(scorching) is composed of two same radical '火'~(fire), which conveys a hot scene. The combination of multiple Chinese characters' glyphs can convey the characteristics of some objects. It is highly consistent with the Pictographic Theory of Chinese characters~\cite{chinese_word} and indicates that only using the glyph features of Chinese characters to represent them has the potential research value. 
\end{CJK}

\begin{table}[t]
\renewcommand\arraystretch{1.1}
    \centering
     \caption{Automatic evaluation results on the validation and test sets of CMSM.}
    \label{tab:exp8}
    \begin{tabular}{lccc}
    \toprule[0.9pt]
    \multicolumn{1}{c}{\textbf{Model}} & \multicolumn{1}{c}{\textbf{LR}} & \multicolumn{1}{c}{\textbf{Dev ACC(\%)}} & \multicolumn{1}{c}{\textbf{Test ACC(\%)}} \\\midrule[0.7pt]
    \# BERT$_{15e}$ & $3e-5$ & $89.41$ & $89.79$ \\
    \# GlyphCRM$_{6e}$ & $3e-5$ & $\textbf{90.71}$ & $\textbf{90.84}$\\
    \bottomrule[0.9pt]
    \end{tabular}
\end{table}

\begin{table}[t]
\renewcommand\arraystretch{1.1}
    \centering
     \caption{Automatic evaluation results on the validation and test sets of low-resource E-commerce Product Review Dataset for Sentiment Analysis task.}
    \label{tab:exp9}
    \begin{tabular}{lccc}
    \toprule[0.9pt]
    \multicolumn{1}{c}{\textbf{Model}} & \multicolumn{1}{c}{\textbf{LR}} & \multicolumn{1}{c}{\textbf{Dev ACC(\%)}} & \multicolumn{1}{c}{\textbf{Test ACC(\%)}} \\\midrule[0.7pt]
    \# BERT$_{15e}$ & $3e-5$ & $79.35$ & $73.61$ \\
    \# GlyphCRM$_{6e}$ & $3e-5$ & $\textbf{81.25}$ & $\textbf{76.07}$\\
    \bottomrule[0.9pt]
    \end{tabular}
\end{table}

\noindent\textbf{Few-shot Learning:} To evaluate the performance of pre-trained models on the low-resource task, we select a few-sample dataset E-commerce Product Review Dataset for Sentiment Analysis\cite{FewCLUE}. Its training, validation, and test set only contain 160, 160, and 610 samples, respectively. The experimental results of Table~\ref{tab:exp9} show that the performance of GlyphCRM outperforms BERT by about 2\% gains in accuracy both on the validation and test sets. It indicates that the generalization and transferability of our model is remarkable, especially on the low-resource task. 

To summarize, GlyphCRM can quickly adapt to various Chinese natural language understanding tasks and achieve promising performances. It is mainly attributed to the glyphs of Chinese characters conveying vivid and significant meanings. Meanwhile, the pre-trained model we design makes full use of the glyph features of Chinese characters and their contextual information. What is important is that our proposed approach can solve the out-of-vocabulary problem, which can be reflected by the improvement on some tests and high transferability on medical fields.

\section{Conclusion}

In this paper, inspired by glyphs of Chinese characters could enhancing the representation of Chinese characters, we propose the Chinese pre-trained representation model named as GlyphCRM, based entirely on glyphs. To verify its effectiveness, we conduct extensive experiments on a wide range of Chinese NLU tasks. The surprising performance of GlyphCRM is that it outperforms previous state-of-the-art model BERT in 9 Chinese NLU tasks. The pre-training process and the fine-tuning results indicate that our model has stronger learning ability, transferability and generalization compared to BERT. It is worth mentioning that a larger pre-trained Chinese representation model GlyphCRM is coming\footnote{We will open the codes and pre-trained checkpoints soon.}.
\bibliographystyle{ACM-Reference-Format}
\bibliography{sample-xelatex}


\begin{thebibliography}{46}


\ifx \showCODEN    \undefined \def \showCODEN     #1{\unskip}     \fi
\ifx \showDOI      \undefined \def \showDOI       #1{#1}\fi
\ifx \showISBNx    \undefined \def \showISBNx     #1{\unskip}     \fi
\ifx \showISBNxiii \undefined \def \showISBNxiii  #1{\unskip}     \fi
\ifx \showISSN     \undefined \def \showISSN      #1{\unskip}     \fi
\ifx \showLCCN     \undefined \def \showLCCN      #1{\unskip}     \fi
\ifx \shownote     \undefined \def \shownote      #1{#1}          \fi
\ifx \showarticletitle \undefined \def \showarticletitle #1{#1}   \fi
\ifx \showURL      \undefined \def \showURL       {\relax}        \fi
\providecommand\bibfield[2]{#2}
\providecommand\bibinfo[2]{#2}
\providecommand\natexlab[1]{#1}
\providecommand\showeprint[2][]{arXiv:#2}

\bibitem[\protect\citeauthoryear{Agarap}{Agarap}{2018}]%
        {relu}
\bibfield{author}{\bibinfo{person}{Abien~Fred Agarap}.}
  \bibinfo{year}{2018}\natexlab{}.
\newblock \showarticletitle{Deep Learning using Rectified Linear Units (ReLU)}.
\newblock \bibinfo{journal}{\emph{CoRR}}  \bibinfo{volume}{abs/1803.08375}
  (\bibinfo{year}{2018}).
\newblock
\showeprint[arxiv]{1803.08375}
\urldef\tempurl%
\url{http://arxiv.org/abs/1803.08375}
\showURL{%
\tempurl}


\bibitem[\protect\citeauthoryear{Ba, Kiros, and Hinton}{Ba
  et~al\mbox{.}}{2016}]%
        {layernorm}
\bibfield{author}{\bibinfo{person}{Jimmy~Lei Ba}, \bibinfo{person}{Jamie~Ryan
  Kiros}, {and} \bibinfo{person}{Geoffrey~E. Hinton}.}
  \bibinfo{year}{2016}\natexlab{}.
\newblock \showarticletitle{Layer Normalization}.
\newblock  (\bibinfo{year}{2016}).
\newblock
\showeprint[arxiv]{1607.06450}~[stat.ML]


\bibitem[\protect\citeauthoryear{benchmark}{benchmark}{2021}]%
        {FewCLUE}
\bibfield{author}{\bibinfo{person}{CLUE benchmark}.}
  \bibinfo{year}{2021}\natexlab{}.
\newblock \bibinfo{title}{FewCLUE:Few-shot learning for Chinese Language
  Understanding Evaluation}.
\newblock
  \bibinfo{howpublished}{\url{https://github.com/CLUEbenchmark/FewCLUE}}.
\newblock


\bibitem[\protect\citeauthoryear{Bordes, Usunier, Chopra, and Weston}{Bordes
  et~al\mbox{.}}{2015}]%
        {questiona2}
\bibfield{author}{\bibinfo{person}{Antoine Bordes}, \bibinfo{person}{Nicolas
  Usunier}, \bibinfo{person}{Sumit Chopra}, {and} \bibinfo{person}{Jason
  Weston}.} \bibinfo{year}{2015}\natexlab{}.
\newblock \showarticletitle{Large-scale Simple Question Answering with Memory
  Networks}.
\newblock \bibinfo{journal}{\emph{CoRR}}  \bibinfo{volume}{abs/1506.02075}
  (\bibinfo{year}{2015}).
\newblock
\showeprint[arxiv]{1506.02075}
\urldef\tempurl%
\url{http://arxiv.org/abs/1506.02075}
\showURL{%
\tempurl}


\bibitem[\protect\citeauthoryear{Brown, Mann, Ryder, Subbiah, Kaplan, Dhariwal,
  Neelakantan, Shyam, Sastry, Askell, Agarwal, Herbert-Voss, Krueger, Henighan,
  Child, Ramesh, Ziegler, Wu, Winter, Hesse, Chen, Sigler, Litwin, Gray, Chess,
  Clark, Berner, McCandlish, Radford, Sutskever, and Amodei}{Brown
  et~al\mbox{.}}{2020}]%
        {gpt_3}
\bibfield{author}{\bibinfo{person}{Tom~B. Brown}, \bibinfo{person}{Benjamin
  Mann}, \bibinfo{person}{Nick Ryder}, \bibinfo{person}{Melanie Subbiah},
  \bibinfo{person}{Jared Kaplan}, \bibinfo{person}{Prafulla Dhariwal},
  \bibinfo{person}{Arvind Neelakantan}, \bibinfo{person}{Pranav Shyam},
  \bibinfo{person}{Girish Sastry}, \bibinfo{person}{Amanda Askell},
  \bibinfo{person}{Sandhini Agarwal}, \bibinfo{person}{Ariel Herbert-Voss},
  \bibinfo{person}{Gretchen Krueger}, \bibinfo{person}{Tom Henighan},
  \bibinfo{person}{Rewon Child}, \bibinfo{person}{Aditya Ramesh},
  \bibinfo{person}{Daniel~M. Ziegler}, \bibinfo{person}{Jeffrey Wu},
  \bibinfo{person}{Clemens Winter}, \bibinfo{person}{Christopher Hesse},
  \bibinfo{person}{Mark Chen}, \bibinfo{person}{Eric Sigler},
  \bibinfo{person}{Mateusz Litwin}, \bibinfo{person}{Scott Gray},
  \bibinfo{person}{Benjamin Chess}, \bibinfo{person}{Jack Clark},
  \bibinfo{person}{Christopher Berner}, \bibinfo{person}{Sam McCandlish},
  \bibinfo{person}{Alec Radford}, \bibinfo{person}{Ilya Sutskever}, {and}
  \bibinfo{person}{Dario Amodei}.} \bibinfo{year}{2020}\natexlab{}.
\newblock \showarticletitle{Language Models are Few-Shot Learners}.
\newblock  (\bibinfo{year}{2020}).
\newblock
\showeprint[arxiv]{2005.14165}~[cs.CL]


\bibitem[\protect\citeauthoryear{Chen, Yu, and Lin}{Chen et~al\mbox{.}}{2020}]%
        {chen-etal-2020-glyph2vec}
\bibfield{author}{\bibinfo{person}{Hong-You Chen}, \bibinfo{person}{Sz-Han Yu},
  {and} \bibinfo{person}{Shou-de Lin}.} \bibinfo{year}{2020}\natexlab{}.
\newblock \showarticletitle{{G}lyph2{V}ec: Learning {C}hinese Out-of-Vocabulary
  Word Embedding from Glyphs}. In \bibinfo{booktitle}{\emph{Proceedings of the
  58th Annual Meeting of the Association for Computational Linguistics}}.
  \bibinfo{publisher}{Association for Computational Linguistics},
  \bibinfo{address}{Online}, \bibinfo{pages}{2865--2871}.
\newblock
\urldef\tempurl%
\url{https://doi.org/10.18653/v1/2020.acl-main.256}
\showDOI{\tempurl}


\bibitem[\protect\citeauthoryear{Cui, Che, Liu, Qin, Wang, and Hu}{Cui
  et~al\mbox{.}}{2020}]%
        {scir_wwm}
\bibfield{author}{\bibinfo{person}{Yiming Cui}, \bibinfo{person}{Wanxiang Che},
  \bibinfo{person}{Ting Liu}, \bibinfo{person}{Bing Qin},
  \bibinfo{person}{Shijin Wang}, {and} \bibinfo{person}{Guoping Hu}.}
  \bibinfo{year}{2020}\natexlab{}.
\newblock \showarticletitle{Revisiting Pre-Trained Models for {C}hinese Natural
  Language Processing}. In \bibinfo{booktitle}{\emph{Proceedings of the 2020
  Conference on Empirical Methods in Natural Language Processing: Findings}}.
  \bibinfo{publisher}{Association for Computational Linguistics},
  \bibinfo{address}{Online}, \bibinfo{pages}{657--668}.
\newblock
\urldef\tempurl%
\url{https://www.aclweb.org/anthology/2020.findings-emnlp.58}
\showURL{%
\tempurl}


\bibitem[\protect\citeauthoryear{Deshmukh and Sethi}{Deshmukh and
  Sethi}{2020}]%
        {ir_bert}
\bibfield{author}{\bibinfo{person}{Anup~Anand Deshmukh} {and}
  \bibinfo{person}{Udhav Sethi}.} \bibinfo{year}{2020}\natexlab{}.
\newblock \showarticletitle{{IR-BERT:} Leveraging {BERT} for Semantic Search in
  Background Linking for News Articles}.
\newblock \bibinfo{journal}{\emph{CoRR}}  \bibinfo{volume}{abs/2007.12603}
  (\bibinfo{year}{2020}).
\newblock
\showeprint[arxiv]{2007.12603}
\urldef\tempurl%
\url{https://arxiv.org/abs/2007.12603}
\showURL{%
\tempurl}


\bibitem[\protect\citeauthoryear{Devlin, Chang, Lee, and Toutanova}{Devlin
  et~al\mbox{.}}{2019}]%
        {devlin-etal-2019-bert}
\bibfield{author}{\bibinfo{person}{Jacob Devlin}, \bibinfo{person}{Ming-Wei
  Chang}, \bibinfo{person}{Kenton Lee}, {and} \bibinfo{person}{Kristina
  Toutanova}.} \bibinfo{year}{2019}\natexlab{}.
\newblock \showarticletitle{{BERT}: Pre-training of Deep Bidirectional
  Transformers for Language Understanding}. In
  \bibinfo{booktitle}{\emph{Proceedings of the 2019 Conference of the North
  {A}merican Chapter of the Association for Computational Linguistics: Human
  Language Technologies, Volume 1 (Long and Short Papers)}}.
  \bibinfo{publisher}{Association for Computational Linguistics},
  \bibinfo{address}{Minneapolis, Minnesota}, \bibinfo{pages}{4171--4186}.
\newblock
\urldef\tempurl%
\url{https://doi.org/10.18653/v1/N19-1423}
\showDOI{\tempurl}


\bibitem[\protect\citeauthoryear{Farmanbar, Van~Ommeren, and Zhao}{Farmanbar
  et~al\mbox{.}}{2020}]%
        {semantic_bert}
\bibfield{author}{\bibinfo{person}{Mojtaba Farmanbar}, \bibinfo{person}{Nikki
  Van~Ommeren}, {and} \bibinfo{person}{Boyang Zhao}.}
  \bibinfo{year}{2020}\natexlab{}.
\newblock \showarticletitle{Semantic search with domain-specific word-embedding
  and production monitoring in Fintech}. In
  \bibinfo{booktitle}{\emph{Proceedings of the 28th International Conference on
  Computational Linguistics: System Demonstrations}}.
  \bibinfo{publisher}{International Committee on Computational Linguistics
  (ICCL)}, \bibinfo{address}{Barcelona, Spain (Online)},
  \bibinfo{pages}{28--33}.
\newblock
\urldef\tempurl%
\url{https://doi.org/10.18653/v1/2020.coling-demos.6}
\showDOI{\tempurl}


\bibitem[\protect\citeauthoryear{Fasel and Luettin}{Fasel and Luettin}{2003}]%
        {perceptron}
\bibfield{author}{\bibinfo{person}{Beat Fasel} {and} \bibinfo{person}{Juergen
  Luettin}.} \bibinfo{year}{2003}\natexlab{}.
\newblock \showarticletitle{Automatic facial expression analysis: a survey}.
\newblock \bibinfo{journal}{\emph{Pattern recognition}} \bibinfo{volume}{36},
  \bibinfo{number}{1} (\bibinfo{year}{2003}), \bibinfo{pages}{259--275}.
\newblock


\bibitem[\protect\citeauthoryear{Gass}{Gass}{2001}]%
        {sentence_m}
\bibfield{author}{\bibinfo{person}{Susan~M. Gass}.}
  \bibinfo{year}{2001}\natexlab{}.
\newblock \showarticletitle{Sentence matching: a re-examination}.
\newblock \bibinfo{journal}{\emph{Second Language Research}}
  \bibinfo{volume}{17}, \bibinfo{number}{4} (\bibinfo{year}{2001}),
  \bibinfo{pages}{421--441}.
\newblock
\urldef\tempurl%
\url{https://doi.org/10.1177/026765830101700407}
\showDOI{\tempurl}
\showeprint{https://doi.org/10.1177/026765830101700407}


\bibitem[\protect\citeauthoryear{Gururangan, Marasovi{\'c}, Swayamdipta, Lo,
  Beltagy, Downey, and Smith}{Gururangan et~al\mbox{.}}{2020}]%
        {dont_stop_pretrain}
\bibfield{author}{\bibinfo{person}{Suchin Gururangan}, \bibinfo{person}{Ana
  Marasovi{\'c}}, \bibinfo{person}{Swabha Swayamdipta}, \bibinfo{person}{Kyle
  Lo}, \bibinfo{person}{Iz Beltagy}, \bibinfo{person}{Doug Downey}, {and}
  \bibinfo{person}{Noah~A. Smith}.} \bibinfo{year}{2020}\natexlab{}.
\newblock \showarticletitle{Don{'}t Stop Pretraining: Adapt Language Models to
  Domains and Tasks}. In \bibinfo{booktitle}{\emph{Proceedings of the 58th
  Annual Meeting of the Association for Computational Linguistics}}.
  \bibinfo{publisher}{Association for Computational Linguistics},
  \bibinfo{address}{Online}, \bibinfo{pages}{8342--8360}.
\newblock
\urldef\tempurl%
\url{https://doi.org/10.18653/v1/2020.acl-main.740}
\showDOI{\tempurl}


\bibitem[\protect\citeauthoryear{He, Zhang, Ren, and Sun}{He
  et~al\mbox{.}}{2016}]%
        {resnet}
\bibfield{author}{\bibinfo{person}{Kaiming He}, \bibinfo{person}{Xiangyu
  Zhang}, \bibinfo{person}{Shaoqing Ren}, {and} \bibinfo{person}{Jian Sun}.}
  \bibinfo{year}{2016}\natexlab{}.
\newblock \showarticletitle{Deep residual learning for image recognition}. In
  \bibinfo{booktitle}{\emph{Proceedings of the IEEE conference on computer
  vision and pattern recognition}}. \bibinfo{pages}{770--778}.
\newblock


\bibitem[\protect\citeauthoryear{Hu, Lu, Li, and Chen}{Hu
  et~al\mbox{.}}{2014}]%
        {NIPS2014_hu}
\bibfield{author}{\bibinfo{person}{Baotian Hu}, \bibinfo{person}{Zhengdong Lu},
  \bibinfo{person}{Hang Li}, {and} \bibinfo{person}{Qingcai Chen}.}
  \bibinfo{year}{2014}\natexlab{}.
\newblock \showarticletitle{Convolutional Neural Network Architectures for
  Matching Natural Language Sentences}. In \bibinfo{booktitle}{\emph{Advances
  in Neural Information Processing Systems}},
  \bibfield{editor}{\bibinfo{person}{Z.~Ghahramani},
  \bibinfo{person}{M.~Welling}, \bibinfo{person}{C.~Cortes},
  \bibinfo{person}{N.~Lawrence}, {and} \bibinfo{person}{K.~Q. Weinberger}}
  (Eds.), Vol.~\bibinfo{volume}{27}. \bibinfo{publisher}{Curran Associates,
  Inc.}
\newblock
\urldef\tempurl%
\url{https://proceedings.neurips.cc/paper/2014/file/b9d487a30398d42ecff55c228ed5652b-Paper.pdf}
\showURL{%
\tempurl}


\bibitem[\protect\citeauthoryear{Jia and Jia}{Jia and Jia}{2005}]%
        {chinese_word}
\bibfield{author}{\bibinfo{person}{Yuxin Jia} {and} \bibinfo{person}{Xuerui
  Jia}.} \bibinfo{year}{2005}\natexlab{}.
\newblock \showarticletitle{Chinese Characters, Chinese Culture and Chinese
  Mind}.
\newblock
\urldef\tempurl%
\url{https://www.semanticscholar.org/paper/Chinese-Characters\%2C-Chinese-Culture-and-Chinese-Jia-Jia/85a2457c5e3101200e2de5f907abf7c79753cd2c}
\showURL{%
\tempurl}


\bibitem[\protect\citeauthoryear{Johnson, Shen, and Liu}{Johnson
  et~al\mbox{.}}{2020}]%
        {nermsra}
\bibfield{author}{\bibinfo{person}{Johnson}, \bibinfo{person}{Shen}, {and}
  \bibinfo{person}{Liu}.} \bibinfo{year}{2020}\natexlab{}.
\newblock \showarticletitle{CWPC-BiAtt: Character–Word–Position Combined
  BiLSTM-Attention for Chinese Named Entity Recognition}.
\newblock \bibinfo{journal}{\emph{Information}}  \bibinfo{volume}{11}
  (\bibinfo{date}{01} \bibinfo{year}{2020}), \bibinfo{pages}{45}.
\newblock
\urldef\tempurl%
\url{https://doi.org/10.3390/info11010045}
\showDOI{\tempurl}


\bibitem[\protect\citeauthoryear{Kingma and Ba}{Kingma and Ba}{2014}]%
        {kingma2014method}
\bibfield{author}{\bibinfo{person}{Diederik~P. Kingma} {and}
  \bibinfo{person}{Jimmy Ba}.} \bibinfo{year}{2014}\natexlab{}.
\newblock \showarticletitle{Adam: A Method for Stochastic Optimization}.
\newblock
\urldef\tempurl%
\url{http://arxiv.org/abs/1412.6980}
\showURL{%
\tempurl}
\newblock
\shownote{Published as a conference paper at the 3rd International Conference
  for Learning Representations, San Diego, 2015.}


\bibitem[\protect\citeauthoryear{Lafferty, McCallum, and Pereira}{Lafferty
  et~al\mbox{.}}{2001}]%
        {laffertyCrf}
\bibfield{author}{\bibinfo{person}{John~D. Lafferty}, \bibinfo{person}{Andrew
  McCallum}, {and} \bibinfo{person}{Fernando C.~N. Pereira}.}
  \bibinfo{year}{2001}\natexlab{}.
\newblock \showarticletitle{Conditional Random Fields: Probabilistic Models for
  Segmenting and Labeling Sequence Data}. In
  \bibinfo{booktitle}{\emph{Proceedings of the Eighteenth International
  Conference on Machine Learning}} \emph{(\bibinfo{series}{ICML '01})}.
  \bibinfo{publisher}{Morgan Kaufmann Publishers Inc.}, \bibinfo{address}{San
  Francisco, CA, USA}, \bibinfo{pages}{282--289}.
\newblock
\showISBNx{1-55860-778-1}
\urldef\tempurl%
\url{http://dl.acm.org/citation.cfm?id=645530.655813}
\showURL{%
\tempurl}


\bibitem[\protect\citeauthoryear{Levow}{Levow}{2006}]%
        {msra}
\bibfield{author}{\bibinfo{person}{Gina-Anne Levow}.}
  \bibinfo{year}{2006}\natexlab{}.
\newblock \showarticletitle{The Third International {C}hinese Language
  Processing Bakeoff: Word Segmentation and Named Entity Recognition}. In
  \bibinfo{booktitle}{\emph{Proceedings of the Fifth {SIGHAN} Workshop on
  {C}hinese Language Processing}}. \bibinfo{publisher}{Association for
  Computational Linguistics}, \bibinfo{address}{Sydney, Australia},
  \bibinfo{pages}{108--117}.
\newblock
\urldef\tempurl%
\url{https://www.aclweb.org/anthology/W06-0115}
\showURL{%
\tempurl}


\bibitem[\protect\citeauthoryear{Lewis, Liu, Goyal, Ghazvininejad, Mohamed,
  Levy, Stoyanov, and Zettlemoyer}{Lewis et~al\mbox{.}}{2020}]%
        {lewis-etal-2020-bart}
\bibfield{author}{\bibinfo{person}{Mike Lewis}, \bibinfo{person}{Yinhan Liu},
  \bibinfo{person}{Naman Goyal}, \bibinfo{person}{Marjan Ghazvininejad},
  \bibinfo{person}{Abdelrahman Mohamed}, \bibinfo{person}{Omer Levy},
  \bibinfo{person}{Veselin Stoyanov}, {and} \bibinfo{person}{Luke
  Zettlemoyer}.} \bibinfo{year}{2020}\natexlab{}.
\newblock \showarticletitle{{BART}: Denoising Sequence-to-Sequence Pre-training
  for Natural Language Generation, Translation, and Comprehension}. In
  \bibinfo{booktitle}{\emph{Proceedings of the 58th Annual Meeting of the
  Association for Computational Linguistics}}. \bibinfo{publisher}{Association
  for Computational Linguistics}, \bibinfo{address}{Online},
  \bibinfo{pages}{7871--7880}.
\newblock
\urldef\tempurl%
\url{https://doi.org/10.18653/v1/2020.acl-main.703}
\showDOI{\tempurl}


\bibitem[\protect\citeauthoryear{Li and Sun}{Li and Sun}{2007}]%
        {thunews}
\bibfield{author}{\bibinfo{person}{Jingyang Li} {and} \bibinfo{person}{Maosong
  Sun}.} \bibinfo{year}{2007}\natexlab{}.
\newblock \showarticletitle{Scalable Term Selection for Text Categorization}.
  In \bibinfo{booktitle}{\emph{Proceedings of the 2007 Joint Conference on
  Empirical Methods in Natural Language Processing and Computational Natural
  Language Learning ({EMNLP}-{C}o{NLL})}}. \bibinfo{publisher}{Association for
  Computational Linguistics}, \bibinfo{address}{Prague, Czech Republic},
  \bibinfo{pages}{774--782}.
\newblock
\urldef\tempurl%
\url{https://www.aclweb.org/anthology/D07-1081}
\showURL{%
\tempurl}


\bibitem[\protect\citeauthoryear{Li, Yatskar, Yin, Hsieh, and Chang}{Li
  et~al\mbox{.}}{2019}]%
        {visualbert}
\bibfield{author}{\bibinfo{person}{Liunian~Harold Li}, \bibinfo{person}{Mark
  Yatskar}, \bibinfo{person}{Da Yin}, \bibinfo{person}{Cho{-}Jui Hsieh}, {and}
  \bibinfo{person}{Kai{-}Wei Chang}.} \bibinfo{year}{2019}\natexlab{}.
\newblock \showarticletitle{VisualBERT: {A} Simple and Performant Baseline for
  Vision and Language}.
\newblock \bibinfo{journal}{\emph{CoRR}}  \bibinfo{volume}{abs/1908.03557}
  (\bibinfo{year}{2019}).
\newblock
\showeprint[arxiv]{1908.03557}
\urldef\tempurl%
\url{http://arxiv.org/abs/1908.03557}
\showURL{%
\tempurl}


\bibitem[\protect\citeauthoryear{Liu, Ott, Goyal, Du, Joshi, Chen, Levy, Lewis,
  Zettlemoyer, and Stoyanov}{Liu et~al\mbox{.}}{2019}]%
        {liu2019roberta}
\bibfield{author}{\bibinfo{person}{Yinhan Liu}, \bibinfo{person}{Myle Ott},
  \bibinfo{person}{Naman Goyal}, \bibinfo{person}{Jingfei Du},
  \bibinfo{person}{Mandar Joshi}, \bibinfo{person}{Danqi Chen},
  \bibinfo{person}{Omer Levy}, \bibinfo{person}{Mike Lewis},
  \bibinfo{person}{Luke Zettlemoyer}, {and} \bibinfo{person}{Veselin
  Stoyanov}.} \bibinfo{year}{2019}\natexlab{}.
\newblock \bibinfo{title}{RoBERTa: A Robustly Optimized BERT Pretraining
  Approach}.
\newblock
\newblock
\urldef\tempurl%
\url{http://arxiv.org/abs/1907.11692}
\showURL{%
\tempurl}
\newblock
\shownote{cite arxiv:1907.11692.}


\bibitem[\protect\citeauthoryear{Martin, Muller, Ortiz~Su{\'a}rez, Dupont,
  Romary, de~la Clergerie, Seddah, and Sagot}{Martin et~al\mbox{.}}{2020}]%
        {martin-etal-2020-french_bert}
\bibfield{author}{\bibinfo{person}{Louis Martin}, \bibinfo{person}{Benjamin
  Muller}, \bibinfo{person}{Pedro~Javier Ortiz~Su{\'a}rez},
  \bibinfo{person}{Yoann Dupont}, \bibinfo{person}{Laurent Romary},
  \bibinfo{person}{{\'E}ric de~la Clergerie}, \bibinfo{person}{Djam{\'e}
  Seddah}, {and} \bibinfo{person}{Beno{\^\i}t Sagot}.}
  \bibinfo{year}{2020}\natexlab{}.
\newblock \showarticletitle{{C}amem{BERT}: a Tasty {F}rench Language Model}. In
  \bibinfo{booktitle}{\emph{Proceedings of the 58th Annual Meeting of the
  Association for Computational Linguistics}}. \bibinfo{publisher}{Association
  for Computational Linguistics}, \bibinfo{address}{Online},
  \bibinfo{pages}{7203--7219}.
\newblock
\urldef\tempurl%
\url{https://doi.org/10.18653/v1/2020.acl-main.645}
\showDOI{\tempurl}


\bibitem[\protect\citeauthoryear{Meng, Wu, Wang, Li, Nie, Yin, Li, Han, Sun,
  and Li}{Meng et~al\mbox{.}}{2019}]%
        {glyce}
\bibfield{author}{\bibinfo{person}{Yuxian Meng}, \bibinfo{person}{Wei Wu},
  \bibinfo{person}{Fei Wang}, \bibinfo{person}{Xiaoya Li},
  \bibinfo{person}{Ping Nie}, \bibinfo{person}{Fan Yin}, \bibinfo{person}{Muyu
  Li}, \bibinfo{person}{Qinghong Han}, \bibinfo{person}{Xiaofei Sun}, {and}
  \bibinfo{person}{Jiwei Li}.} \bibinfo{year}{2019}\natexlab{}.
\newblock \showarticletitle{Glyce: Glyph-vectors for Chinese Character
  Representations}. In \bibinfo{booktitle}{\emph{Advances in Neural Information
  Processing Systems}}, \bibfield{editor}{\bibinfo{person}{H.~Wallach},
  \bibinfo{person}{H.~Larochelle}, \bibinfo{person}{A.~Beygelzimer},
  \bibinfo{person}{F.~d\textquotesingle Alch\'{e}-Buc},
  \bibinfo{person}{E.~Fox}, {and} \bibinfo{person}{R.~Garnett}} (Eds.),
  Vol.~\bibinfo{volume}{32}. \bibinfo{publisher}{Curran Associates, Inc.}
\newblock
\urldef\tempurl%
\url{https://proceedings.neurips.cc/paper/2019/file/452bf208bf901322968557227b8f6efe-Paper.pdf}
\showURL{%
\tempurl}


\bibitem[\protect\citeauthoryear{Mikolov, Sutskever, Chen, Corrado, and
  Dean}{Mikolov et~al\mbox{.}}{2013}]%
        {word2vec}
\bibfield{author}{\bibinfo{person}{Tomas Mikolov}, \bibinfo{person}{Ilya
  Sutskever}, \bibinfo{person}{Kai Chen}, \bibinfo{person}{Greg~S Corrado},
  {and} \bibinfo{person}{Jeff Dean}.} \bibinfo{year}{2013}\natexlab{}.
\newblock \showarticletitle{Distributed Representations of Words and Phrases
  and their Compositionality}. In \bibinfo{booktitle}{\emph{Advances in Neural
  Information Processing Systems}}, \bibfield{editor}{\bibinfo{person}{C.~J.~C.
  Burges}, \bibinfo{person}{L.~Bottou}, \bibinfo{person}{M.~Welling},
  \bibinfo{person}{Z.~Ghahramani}, {and} \bibinfo{person}{K.~Q. Weinberger}}
  (Eds.), Vol.~\bibinfo{volume}{26}. \bibinfo{publisher}{Curran Associates,
  Inc.}
\newblock
\urldef\tempurl%
\url{https://proceedings.neurips.cc/paper/2013/file/9aa42b31882ec039965f3c4923ce901b-Paper.pdf}
\showURL{%
\tempurl}


\bibitem[\protect\citeauthoryear{Minaee, Kalchbrenner, Cambria, Nikzad,
  Chenaghlu, and Gao}{Minaee et~al\mbox{.}}{2021}]%
        {text_classification}
\bibfield{author}{\bibinfo{person}{Shervin Minaee}, \bibinfo{person}{Nal
  Kalchbrenner}, \bibinfo{person}{Erik Cambria}, \bibinfo{person}{Narjes
  Nikzad}, \bibinfo{person}{Meysam Chenaghlu}, {and} \bibinfo{person}{Jianfeng
  Gao}.} \bibinfo{year}{2021}\natexlab{}.
\newblock \showarticletitle{Deep Learning--Based Text Classification: A
  Comprehensive Review}.
\newblock  \bibinfo{volume}{54}, \bibinfo{number}{3} (\bibinfo{year}{2021}).
\newblock
\showISSN{0360-0300}
\urldef\tempurl%
\url{https://doi.org/10.1145/3439726}
\showDOI{\tempurl}


\bibitem[\protect\citeauthoryear{Pennington, Socher, and Manning}{Pennington
  et~al\mbox{.}}{2014}]%
        {pennington-etal-2014-glove}
\bibfield{author}{\bibinfo{person}{Jeffrey Pennington},
  \bibinfo{person}{Richard Socher}, {and} \bibinfo{person}{Christopher
  Manning}.} \bibinfo{year}{2014}\natexlab{}.
\newblock \showarticletitle{{G}lo{V}e: Global Vectors for Word Representation}.
  In \bibinfo{booktitle}{\emph{Proceedings of the 2014 Conference on Empirical
  Methods in Natural Language Processing ({EMNLP})}}.
  \bibinfo{publisher}{Association for Computational Linguistics},
  \bibinfo{address}{Doha, Qatar}, \bibinfo{pages}{1532--1543}.
\newblock
\urldef\tempurl%
\url{https://doi.org/10.3115/v1/D14-1162}
\showDOI{\tempurl}


\bibitem[\protect\citeauthoryear{Peters, Ammar, Bhagavatula, and Power}{Peters
  et~al\mbox{.}}{2017}]%
        {elmo}
\bibfield{author}{\bibinfo{person}{Matthew Peters}, \bibinfo{person}{Waleed
  Ammar}, \bibinfo{person}{Chandra Bhagavatula}, {and} \bibinfo{person}{Russell
  Power}.} \bibinfo{year}{2017}\natexlab{}.
\newblock \showarticletitle{Semi-supervised sequence tagging with bidirectional
  language models}. In \bibinfo{booktitle}{\emph{Proceedings of the 55th Annual
  Meeting of the Association for Computational Linguistics (Volume 1: Long
  Papers)}}. \bibinfo{publisher}{Association for Computational Linguistics},
  \bibinfo{address}{Vancouver, Canada}, \bibinfo{pages}{1756--1765}.
\newblock
\urldef\tempurl%
\url{https://doi.org/10.18653/v1/P17-1161}
\showDOI{\tempurl}


\bibitem[\protect\citeauthoryear{Peters, Neumann, Iyyer, Gardner, Clark, Lee,
  and Zettlemoyer}{Peters et~al\mbox{.}}{2018}]%
        {elmo_b}
\bibfield{author}{\bibinfo{person}{Matthew Peters}, \bibinfo{person}{Mark
  Neumann}, \bibinfo{person}{Mohit Iyyer}, \bibinfo{person}{Matt Gardner},
  \bibinfo{person}{Christopher Clark}, \bibinfo{person}{Kenton Lee}, {and}
  \bibinfo{person}{Luke Zettlemoyer}.} \bibinfo{year}{2018}\natexlab{}.
\newblock \showarticletitle{Deep Contextualized Word Representations}. In
  \bibinfo{booktitle}{\emph{Proceedings of the 2018 Conference of the North
  {A}merican Chapter of the Association for Computational Linguistics: Human
  Language Technologies, Volume 1 (Long Papers)}}.
  \bibinfo{publisher}{Association for Computational Linguistics},
  \bibinfo{address}{New Orleans, Louisiana}, \bibinfo{pages}{2227--2237}.
\newblock
\urldef\tempurl%
\url{https://doi.org/10.18653/v1/N18-1202}
\showDOI{\tempurl}


\bibitem[\protect\citeauthoryear{Radford and Sutskever}{Radford and
  Sutskever}{2018}]%
        {gpt}
\bibfield{author}{\bibinfo{person}{Alec Radford} {and} \bibinfo{person}{Ilya
  Sutskever}.} \bibinfo{year}{2018}\natexlab{}.
\newblock \showarticletitle{Improving Language Understanding by Generative
  Pre-Training}. In \bibinfo{booktitle}{\emph{arxiv}}.
\newblock


\bibitem[\protect\citeauthoryear{Radford, Wu, Child, Luan, Amodei, and
  Sutskever}{Radford et~al\mbox{.}}{2019}]%
        {radford2019language_gpt_2}
\bibfield{author}{\bibinfo{person}{Alec Radford}, \bibinfo{person}{Jeff Wu},
  \bibinfo{person}{Rewon Child}, \bibinfo{person}{David Luan},
  \bibinfo{person}{Dario Amodei}, {and} \bibinfo{person}{Ilya Sutskever}.}
  \bibinfo{year}{2019}\natexlab{}.
\newblock \showarticletitle{Language Models are Unsupervised Multitask
  Learners}.
\newblock  (\bibinfo{year}{2019}).
\newblock


\bibitem[\protect\citeauthoryear{Raffel, Shazeer, Roberts, Lee, Narang, Matena,
  Zhou, Li, and Liu}{Raffel et~al\mbox{.}}{2019}]%
        {T5}
\bibfield{author}{\bibinfo{person}{Colin Raffel}, \bibinfo{person}{Noam
  Shazeer}, \bibinfo{person}{Adam Roberts}, \bibinfo{person}{Katherine Lee},
  \bibinfo{person}{Sharan Narang}, \bibinfo{person}{Michael Matena},
  \bibinfo{person}{Yanqi Zhou}, \bibinfo{person}{Wei Li}, {and}
  \bibinfo{person}{Peter~J. Liu}.} \bibinfo{year}{2019}\natexlab{}.
\newblock \showarticletitle{Exploring the Limits of Transfer Learning with a
  Unified Text-to-Text Transformer}.
\newblock \bibinfo{journal}{\emph{CoRR}}  \bibinfo{volume}{abs/1910.10683}.
\newblock
\showeprint[arxiv]{1910.10683}
\urldef\tempurl%
\url{http://arxiv.org/abs/1910.10683}
\showURL{%
\tempurl}


\bibitem[\protect\citeauthoryear{Reimers and Gurevych}{Reimers and
  Gurevych}{2019}]%
        {reimers-gurevych-2019-sentence_bert}
\bibfield{author}{\bibinfo{person}{Nils Reimers} {and} \bibinfo{person}{Iryna
  Gurevych}.} \bibinfo{year}{2019}\natexlab{}.
\newblock \showarticletitle{Sentence-{BERT}: Sentence Embeddings using
  {S}iamese {BERT}-Networks}. In \bibinfo{booktitle}{\emph{Proceedings of the
  2019 Conference on Empirical Methods in Natural Language Processing and the
  9th International Joint Conference on Natural Language Processing
  (EMNLP-IJCNLP)}}. \bibinfo{publisher}{Association for Computational
  Linguistics}, \bibinfo{address}{Hong Kong, China},
  \bibinfo{pages}{3982--3992}.
\newblock
\urldef\tempurl%
\url{https://doi.org/10.18653/v1/D19-1410}
\showDOI{\tempurl}


\bibitem[\protect\citeauthoryear{Sahlgren}{Sahlgren}{2008}]%
        {sahlgren2008distributional}
\bibfield{author}{\bibinfo{person}{Magnus Sahlgren}.}
  \bibinfo{year}{2008}\natexlab{}.
\newblock \showarticletitle{The distributional hypothesis}.
\newblock \bibinfo{journal}{\emph{Italian Journal of Disability Studies}}
  \bibinfo{volume}{20} (\bibinfo{year}{2008}), \bibinfo{pages}{33--53}.
\newblock


\bibitem[\protect\citeauthoryear{Sergio and Lee}{Sergio and Lee}{2020}]%
        {debert}
\bibfield{author}{\bibinfo{person}{Gwenaelle~Cunha Sergio} {and}
  \bibinfo{person}{Minho Lee}.} \bibinfo{year}{2020}\natexlab{}.
\newblock \showarticletitle{Stacked DeBERT: All Attention in Incomplete Data
  for Text Classification}.
\newblock \bibinfo{journal}{\emph{CoRR}}  \bibinfo{volume}{abs/2001.00137}
  (\bibinfo{year}{2020}).
\newblock
\showeprint[arxiv]{2001.00137}
\urldef\tempurl%
\url{http://arxiv.org/abs/2001.00137}
\showURL{%
\tempurl}


\bibitem[\protect\citeauthoryear{Su and Lee}{Su and Lee}{2017}]%
        {learning_glyph}
\bibfield{author}{\bibinfo{person}{Tzu-Ray Su} {and} \bibinfo{person}{Hung-Yi
  Lee}.} \bibinfo{year}{2017}\natexlab{}.
\newblock \showarticletitle{Learning {C}hinese Word Representations From Glyphs
  Of Characters}. In \bibinfo{booktitle}{\emph{Proceedings of the 2017
  Conference on Empirical Methods in Natural Language Processing}}.
  \bibinfo{publisher}{Association for Computational Linguistics},
  \bibinfo{address}{Copenhagen, Denmark}, \bibinfo{pages}{264--273}.
\newblock
\urldef\tempurl%
\url{https://doi.org/10.18653/v1/D17-1025}
\showDOI{\tempurl}


\bibitem[\protect\citeauthoryear{Tan and Bansal}{Tan and Bansal}{2019}]%
        {tan-bansal-2019-lxmert}
\bibfield{author}{\bibinfo{person}{Hao Tan} {and} \bibinfo{person}{Mohit
  Bansal}.} \bibinfo{year}{2019}\natexlab{}.
\newblock \showarticletitle{{LXMERT}: Learning Cross-Modality Encoder
  Representations from Transformers}. In \bibinfo{booktitle}{\emph{Proceedings
  of the 2019 Conference on Empirical Methods in Natural Language Processing
  and the 9th International Joint Conference on Natural Language Processing
  (EMNLP-IJCNLP)}}. \bibinfo{publisher}{Association for Computational
  Linguistics}, \bibinfo{address}{Hong Kong, China},
  \bibinfo{pages}{5100--5111}.
\newblock
\urldef\tempurl%
\url{https://doi.org/10.18653/v1/D19-1514}
\showDOI{\tempurl}


\bibitem[\protect\citeauthoryear{Tan and Zhang}{Tan and Zhang}{2008}]%
        {chnSentiCorp}
\bibfield{author}{\bibinfo{person}{Songbo Tan} {and} \bibinfo{person}{Jin
  Zhang}.} \bibinfo{year}{2008}\natexlab{}.
\newblock \showarticletitle{An empirical study of sentiment analysis for
  chinese documents}.
\newblock \bibinfo{journal}{\emph{Expert Systems with Applications}}
  \bibinfo{volume}{34}, \bibinfo{number}{4} (\bibinfo{year}{2008}),
  \bibinfo{pages}{2622--2629}.
\newblock
\showISSN{0957-4174}
\urldef\tempurl%
\url{https://doi.org/10.1016/j.eswa.2007.05.028}
\showDOI{\tempurl}


\bibitem[\protect\citeauthoryear{Vaswani, Shazeer, Parmar, Uszkoreit, Jones,
  Gomez, Kaiser, and Polosukhin}{Vaswani et~al\mbox{.}}{2017}]%
        {attention}
\bibfield{author}{\bibinfo{person}{Ashish Vaswani}, \bibinfo{person}{Noam
  Shazeer}, \bibinfo{person}{Niki Parmar}, \bibinfo{person}{Jakob Uszkoreit},
  \bibinfo{person}{Llion Jones}, \bibinfo{person}{Aidan~N Gomez},
  \bibinfo{person}{\L~ukasz Kaiser}, {and} \bibinfo{person}{Illia Polosukhin}.}
  \bibinfo{year}{2017}\natexlab{}.
\newblock \showarticletitle{Attention is All you Need}. In
  \bibinfo{booktitle}{\emph{Advances in Neural Information Processing
  Systems}}, \bibfield{editor}{\bibinfo{person}{I.~Guyon},
  \bibinfo{person}{U.~V. Luxburg}, \bibinfo{person}{S.~Bengio},
  \bibinfo{person}{H.~Wallach}, \bibinfo{person}{R.~Fergus},
  \bibinfo{person}{S.~Vishwanathan}, {and} \bibinfo{person}{R.~Garnett}}
  (Eds.), Vol.~\bibinfo{volume}{30}. \bibinfo{publisher}{Curran Associates,
  Inc.}
\newblock
\urldef\tempurl%
\url{https://proceedings.neurips.cc/paper/2017/file/3f5ee243547dee91fbd053c1c4a845aa-Paper.pdf}
\showURL{%
\tempurl}


\bibitem[\protect\citeauthoryear{Wang, Yang, Wei, Chang, and Zhou}{Wang
  et~al\mbox{.}}{2017}]%
        {wang-etal-2017-questiona}
\bibfield{author}{\bibinfo{person}{Wenhui Wang}, \bibinfo{person}{Nan Yang},
  \bibinfo{person}{Furu Wei}, \bibinfo{person}{Baobao Chang}, {and}
  \bibinfo{person}{Ming Zhou}.} \bibinfo{year}{2017}\natexlab{}.
\newblock \showarticletitle{Gated Self-Matching Networks for Reading
  Comprehension and Question Answering}. In
  \bibinfo{booktitle}{\emph{Proceedings of the 55th Annual Meeting of the
  Association for Computational Linguistics (Volume 1: Long Papers)}}.
  \bibinfo{publisher}{Association for Computational Linguistics},
  \bibinfo{address}{Vancouver, Canada}, \bibinfo{pages}{189--198}.
\newblock
\urldef\tempurl%
\url{https://doi.org/10.18653/v1/P17-1018}
\showDOI{\tempurl}


\bibitem[\protect\citeauthoryear{Xue, Constant, Roberts, Kale, Al{-}Rfou,
  Siddhant, Barua, and Raffel}{Xue et~al\mbox{.}}{2020}]%
        {mt5}
\bibfield{author}{\bibinfo{person}{Linting Xue}, \bibinfo{person}{Noah
  Constant}, \bibinfo{person}{Adam Roberts}, \bibinfo{person}{Mihir Kale},
  \bibinfo{person}{Rami Al{-}Rfou}, \bibinfo{person}{Aditya Siddhant},
  \bibinfo{person}{Aditya Barua}, {and} \bibinfo{person}{Colin Raffel}.}
  \bibinfo{year}{2020}\natexlab{}.
\newblock \showarticletitle{mT5: {A} massively multilingual pre-trained
  text-to-text transformer}.
\newblock \bibinfo{journal}{\emph{CoRR}}  \bibinfo{volume}{abs/2010.11934}
  (\bibinfo{year}{2020}).
\newblock
\showeprint[arxiv]{2010.11934}
\urldef\tempurl%
\url{https://arxiv.org/abs/2010.11934}
\showURL{%
\tempurl}


\bibitem[\protect\citeauthoryear{Yang, Dai, Yang, Carbonell, Salakhutdinov, and
  Le}{Yang et~al\mbox{.}}{2019}]%
        {xlnet}
\bibfield{author}{\bibinfo{person}{Zhilin Yang}, \bibinfo{person}{Zihang Dai},
  \bibinfo{person}{Yiming Yang}, \bibinfo{person}{Jaime Carbonell},
  \bibinfo{person}{Russ~R Salakhutdinov}, {and} \bibinfo{person}{Quoc~V Le}.}
  \bibinfo{year}{2019}\natexlab{}.
\newblock \showarticletitle{XLNet: Generalized Autoregressive Pretraining for
  Language Understanding}. In \bibinfo{booktitle}{\emph{Advances in Neural
  Information Processing Systems}},
  \bibfield{editor}{\bibinfo{person}{H.~Wallach},
  \bibinfo{person}{H.~Larochelle}, \bibinfo{person}{A.~Beygelzimer},
  \bibinfo{person}{F.~d\textquotesingle Alch\'{e}-Buc},
  \bibinfo{person}{E.~Fox}, {and} \bibinfo{person}{R.~Garnett}} (Eds.),
  Vol.~\bibinfo{volume}{32}. \bibinfo{publisher}{Curran Associates, Inc.}
\newblock
\urldef\tempurl%
\url{https://proceedings.neurips.cc/paper/2019/file/dc6a7e655d7e5840e66733e9ee67cc69-Paper.pdf}
\showURL{%
\tempurl}


\bibitem[\protect\citeauthoryear{Yu, Tang, Yin, Sun, Tian, Wu, and Wang}{Yu
  et~al\mbox{.}}{2020}]%
        {ernie_vil}
\bibfield{author}{\bibinfo{person}{Fei Yu}, \bibinfo{person}{Jiji Tang},
  \bibinfo{person}{Weichong Yin}, \bibinfo{person}{Yu Sun},
  \bibinfo{person}{Hao Tian}, \bibinfo{person}{Hua Wu}, {and}
  \bibinfo{person}{Haifeng Wang}.} \bibinfo{year}{2020}\natexlab{}.
\newblock \showarticletitle{ERNIE-ViL: Knowledge Enhanced Vision-Language
  Representations Through Scene Graph}.
\newblock \bibinfo{journal}{\emph{CoRR}}  \bibinfo{volume}{abs/2006.16934}
  (\bibinfo{year}{2020}).
\newblock
\showeprint[arxiv]{2006.16934}
\urldef\tempurl%
\url{https://arxiv.org/abs/2006.16934}
\showURL{%
\tempurl}


\bibitem[\protect\citeauthoryear{Zhang, Han, Liu, Jiang, Sun, and Liu}{Zhang
  et~al\mbox{.}}{2019}]%
        {zhang-etal-2019-ernie}
\bibfield{author}{\bibinfo{person}{Zhengyan Zhang}, \bibinfo{person}{Xu Han},
  \bibinfo{person}{Zhiyuan Liu}, \bibinfo{person}{Xin Jiang},
  \bibinfo{person}{Maosong Sun}, {and} \bibinfo{person}{Qun Liu}.}
  \bibinfo{year}{2019}\natexlab{}.
\newblock \showarticletitle{{ERNIE}: Enhanced Language Representation with
  Informative Entities}. In \bibinfo{booktitle}{\emph{Proceedings of the 57th
  Annual Meeting of the Association for Computational Linguistics}}.
  \bibinfo{publisher}{Association for Computational Linguistics},
  \bibinfo{address}{Florence, Italy}, \bibinfo{pages}{1441--1451}.
\newblock
\urldef\tempurl%
\url{https://doi.org/10.18653/v1/P19-1139}
\showDOI{\tempurl}


\end{thebibliography}

\end{document}